\documentclass{article}
\usepackage{graphicx} 
\usepackage{amsmath}
\usepackage{amsfonts}
\usepackage[table, svgnames, dvipsnames]{xcolor}
\definecolor{RoyalBlue}{HTML}{8B008B}
\usepackage[colorlinks=true, linkcolor=RoyalBlue, citecolor=RoyalBlue, urlcolor=RoyalBlue]{hyperref}
\usepackage[round]{natbib}
\usepackage{multirow}
\usepackage{tabularx}
\usepackage{array}
\usepackage{colortbl}
\usepackage{booktabs}
\usepackage{siunitx}
\usepackage{comment}
\usepackage[ruled,vlined]{algorithm2e}

\usepackage{caption}
\newcommand{\EX}{\mathbb{E}}
\DeclareMathOperator*{\argmin}{arg\,min}
\DeclareMathOperator*{\argmax}{arg\,max}

\newcommand*{\belowrulesepcolor}[1]{%
  \noalign{%
    \kern-\belowrulesep 
    \begingroup 
      \color{#1}%
      \hrule height\belowrulesep 
    \endgroup 
  }%
} 
\newcommand*{\aboverulesepcolor}[1]{%
  \noalign{%
    \begingroup 
      \color{#1}%
      \hrule height\aboverulesep 
    \endgroup 
    \kern-\aboverulesep 
  }%
}

\allowdisplaybreaks

\title{Complex variational autoencoders admit Kähler structure}
\author{Andrew Gracyk\thanks{{\normalsize Purdue University; agracyk@purdue.edu}}}
\date{}

\begin{document}

\maketitle

\abstract{It has been discovered that latent-Euclidean variational autoencoders (VAEs) admit, in various capacities, Riemannian structure. We adapt these arguments but for complex VAEs with a complex latent stage. We show that complex VAEs reveal to some level Kähler geometric structure. Our methods will be tailored for decoder geometry. We derive the Fisher information metric in the complex case under a latent complex Gaussian with trivial relation matrix. It is well known from statistical information theory that the Fisher information coincides with the Hessian of the Kullback-Leibler (KL) divergence. Thus, the metric Kähler potential relation is exactly achieved under relative entropy. We propose a Kähler potential derivative of complex Gaussian mixtures that acts as a rough proxy to the Fisher information metric while still being faithful to the underlying Kähler geometry. Computation of the metric via this potential is efficient, and through our potential, valid as a plurisubharmonic (PSH) function, large scale computational burden of automatic differentiation is displaced to small scale. Our methods leverage the law of total covariance to bridge behavior between our potential and the Fisher metric. We show that we can regularize the latent space with decoder geometry, and that we can sample in accordance with a weighted complex volume element. We demonstrate these strategies, at the exchange of sample variation, yield consistently smoother representations and fewer semantic outliers.}


\tableofcontents

\section{Introduction}

We study the auxiliary qualities of the variational autoencoder (VAE) via the complex geometric perspective. The Euclidean VAE is a foundational machine learning archetype, and from it has stemmed the complex analog \cite{xie2023complexrecurrentvariationalautoencoder}, although this technology has less establishment in literature. We apply the overarching arguments presented in for imaginary latent data. In particular, \cite{chadebec2022geometricperspectivevariationalautoencoders} presents justification that VAEs admit Riemannian structure in the Euclidean-immersed case. We reframe this argument in a new light, and so we argue imaginary latent data holds the same features but through Kähler geometry.

\vspace{2mm}

\noindent The overarching effects of geometric features for latent variational tasks \cite{kingma2022autoencodingvariationalbayes} has been studied in tasks for external imposition, such as in \cite{lopez2025gdvaesgeometricdynamicvariational} and \cite{palma2025enforcinglatenteuclideangeometry}. In this work, we study the effects of geometric feature for those that form independent of outside force, and rather those that arise natural in training, consistent with the work of \cite{chadebec2022geometricperspectivevariationalautoencoders} but in the context of complex geometry. We refer to \cite{lobashev2025hessiangeometrylatentspace} \cite{zeng2025gmaplatentgeometricmappinglatent} \cite{lee2025geometrypreservingencoderdecoderlatentgenerative} \cite{chadebec2020geometryawarehamiltonianvariationalautoencoder} \cite{kouzelis2025eqvaeequivarianceregularizedlatent} for other developments regarding latent geometry, primarily for variational inference, thus geometric quality in inference and generative tasks is a widely studied problem. In particular, latent geometry certainly develops, but the natural geometry is only part of the picture, i.e. there exist geometries outside of those that can form in the model which can be enforced independently \cite{DBLP:journals/pami/DuqueMWM23}. We will reconcile this paradigm in our arguments via our computational results.

\vspace{2mm}

\noindent Our experiments will be presented in such a way that we attempt to argue the decoder structure and anomaly detection are intertwined. \cite{bergamin2022novelapplicationsvaebasedanomaly} \cite{chauhan2022robustoutlierdetectiondebiasing} \cite{eduardo2020robustvariationalautoencodersoutlier} are a collection of works that implore outlier detection in variational contexts. Outliers in VAEs primarily contain origins from ill-conditioning in the decoder map. This is a known phenomenom to some degree, such as when traversing a class boundary \cite{chen2020learningflatlatentmanifolds}, also briefly discussed in \cite{ramanaik2023ensuringtopologicaldatastructurepreservation} via unbounded Lipschitz constant of the decoder, and moreover studied in \cite{kumar2020implicitregularizationbetavaes}. Thus, drawing from a smooth, desirable Hermitian metric will ramify an amicable sampling.

\section{Notation and conventions}

We will use notation $\dagger$ to denote a Hermitian transpose $X_{ij}^{\dagger} = \overline{X}_{ji}$, $\overline{z}$ to denote the complex conjugate of $z$. $h$ will denote some notion, since we will subsequently use indices and variants, of a Hermitian metric. $z$ will denote a point in the latent space. We will use notation
\begin{align}
\partial \overline{\partial} : \Omega^{0,0} \rightarrow \Omega^{1,1}
\end{align}
to denote the complex Hessian with Dolbeault operators, i.e. it is a differential operator on smooth sections. We will moreover denote
\begin{align}
(\partial \overline{\partial} f)_{\alpha \overline{\beta}} = \partial_{\alpha} \partial_{\overline{\beta}} f 
\end{align}
to be the complex Hessian, since $\partial \overline{\partial} f $ often denotes a $(1,1)$-form. We will use $\Psi$ to denote the (positive) quadratic form in the exponent of a Gaussian density. In general, our notation is mostly standard. For the remainder of our work, we will assume the mixed derivative
\begin{align}
\frac{\partial^2}{\partial z_{\alpha} \partial z_{\overline{\beta}}} \mu(z) = 0, \ \ \ \text{where} \ \ \ \frac{\partial}{\partial z} = \frac{1}{2} \Big( \frac{d}{dx} - i \frac{d}{dy} \Big),\frac{\partial}{\partial \overline{z}} = \frac{1}{2} \Big( \frac{d}{dx} + i \frac{d}{dy} \Big)  
\end{align}
are the Wirtinger derivatives and $\mu$ is a mean vector (we are not saying all mixed Hessians vanish). We will show this is empirically supported in Appendix \ref{app:additional_figures}, Figures \ref{fig:hii_hist}, \ref{fig:first_deriv_hist}. In particular, we find our architectures work well with this criterion. We specifically do not restrict our decoder to be holomorphic since it is well known a real-valued function that is holomorphic over a connected and open domain is constant due to the Cauchy-Riemann equations. In practice, we will compute Wirtinger derivatives using real-valued surrogates in our VAEs for simplicity reasons. For complex integration, we will use the fact that complex integration has an equivalent form over $\mathbb{R}^{p}$
\begin{align}
\int_{\mathbb{C}^d} \Gamma d\lambda_{2d} = \int_{\mathbb{R}^{2d}} \Gamma(x+iy) d\text{Vol}_{\text{Euclidean}} .
\end{align}

\section{The complex variational autoencoder (VAE)}

The complex VAE archetype has the same fundamental underpinning as the typical VAE. We will treat our real-valued data as in complex space. Because it is learned, it is not fully real exactly but under approximation until the real part is taken. Let our pixel data be such that
\begin{align}
& x \in \mathbb{C}^{n} = \{ a + b i : a,b \in \mathbb{R}^n, i = \sqrt{-1} \} \cong \mathbb{R}^{2n}
\\
&  \EX_{q} [ \text{decoder}(z) ] \in \mathbb{C}^{n}, \ \ \ \Big| \text{Im} \Big( \EX_{q} [ \text{decoder}(z) ]  \Big) \Big| < \epsilon  ,
\end{align}
being such that the imaginary pixel data via the learned relation is arbitrarily small, but not nonzero until the real part is taken. Here, $\epsilon \in \mathbb{R}^+$ is a number dependent on training iteration.

\vspace{2mm} 

\noindent As with the typical VAE, we follow suit of \cite{xie2023complexrecurrentvariationalautoencoder}, the objective is
\begin{align}
\EX_{x \sim p_{\text{training}}} [ \log p(x) ] & \geq \EX_{x \sim p_{\text{training}}} [  \EX_{q_{\phi}(z|x)} [ \log p_{\theta}(x|z) ] - \text{KL}( q_{\phi}(z|x) \parallel p(z) ) ] 
\\
& \propto \EX_{x \sim p_{\text{training}}} [  [|| x - g_{\theta}(z) ||_2^2 - \text{KL}( q_{\phi}(z|x) \parallel p(z) ) ]:= \text{VAE loss} .
\end{align}
The outside expectation is less conventional in literature, but it refers to an (empirical) average over training data. Now our variables are complex-valued. Our neural networks will be such that
\begin{align}
f_{\phi} : \mathbb{C}^n \rightarrow (\mu,\sigma,\delta) \in \mathbb{C}^{d} \times \mathbb{R}^d \times \mathbb{R}^d, \ \ \ \ \ g_{\theta} : \mathbb{C}^d \rightarrow \mathbb{C}^n .
\end{align}
Continuing to follow \cite{xie2023complexrecurrentvariationalautoencoder}, we take
\begin{align}
& \tilde{z}_i = z_i + \psi_{\text{Re}} \odot \epsilon_{\text{Re}} + \psi_{\text{Im}} \odot \epsilon_{\text{Im}}
\\
&  \psi_{\text{Re}} = \frac{\sigma + \delta}{2 \sigma + 2 \text{Re}(\delta) },   \psi_{\text{Im}} = i \frac{\sqrt{\sigma^2 -  |\delta|^2}}{2 \sigma + 2 \text{Re}(\delta) } .
\end{align}
Note that we have the closed form
\begin{align}
\text{KL}( q_{\phi}(z|x) \parallel p(z) ) = \mu^{\dagger} \mu  + || \sigma - 1 - \frac{1}{2} \log( \sigma^2 - |\delta|^2 ) ||_1 .
\end{align}

\section{Euclidean VAEs unveil Riemannian structure}

The standard VAE is taken such that
\begin{align}
q_{\phi}(z | x) \sim \mathcal{N}(\mu(x_i), \Sigma(x_i) ) .
\end{align}
The Riemannian metric
\begin{align}
g_z : T_z M \times T_z M \rightarrow \mathbb{R}, g(Y,Z) = g_{ij} Y^i Z^j, Y = Y^i \frac{\partial}{\partial y^i}, Z = Z^j \frac{\partial}{\partial z^j}  ,
\end{align}
or equivalently
\begin{align}
g = g_{ij} dz^i \otimes dz^j ,
\end{align}
can be viewed such that
\begin{align}
g_{\EX[f_{\phi}(x_i)]} = \Sigma^{-1}(x_i) = \nabla_{\mu}^2 ( - \log p (x | \mu, \Sigma) )  .
\end{align}
We will operate so that $\Sigma$ is positive definite, thus $\Sigma^{-1}$ and $g$ are too. Further using the framework of \cite{chadebec2022geometricperspectivevariationalautoencoders}, they propose a smooth continuous Riemannian metric on the entirety of the latent space using kernels and regularization of $||z||$ with temperature. This temperature-normed term will be a motif for us, and it will appear later in our regularization techniques. This smooth metric can be extended under Taylor expansions
\begin{align}
& G(z) \approx \Sigma^{-1}(x_i) + \sum_{j\neq i} \Sigma^{-1}(x_j) \cdot \Omega_j(\mu(x_i)) + \Sigma^{-1}(x_i) \cdot J_{\Omega_i}(\mu(x_i)) \cdot (z - \mu(x_i)) \\
& \Omega_i(z) = \exp \Bigg\{ - \frac{(z-\mu(x_i))^T \Sigma^{-1}(x_i) (z - \mu(x_i))}{\text{constant}^2 } \Bigg\}.
\end{align}
The Riemannian metric of the decoder differs to the sampling of the posterior. It is known $G(z) = J_g^T(z) J_g(z)$ on the latent spaces of generative models via the decoder. This work proposes sampling according to the Radon-Nikodym derivative
\begin{gather}
\label{eqn:uniform_metric}
\frac{d\mathbb{P}_G}{d\lambda^d}(z) = \sqrt{\text{det}(G(z))} \Bigg/ \int_{\mathbb{R}^d} \sqrt{\text{det}(G(z))} dz 
\\
\text{or similarly, for us,}
\\
\frac{d\mathbb{P}_h}{d\lambda^{2d}}(z) = \text{det}(h_{\alpha \overline{\beta}}(z)) \Bigg/ \int_{\mathbb{C}^d} \text{det}(h_{\alpha \overline{\beta}}(z)) d\lambda^{2d}
\end{gather}
(under classical regularity conditions such as absolute continuity w.r.t. alternative measures, and slight abuse of notation on the subscripts). Thus the sampling is done with the constructed $G$ (or a valid surrogate or replacement). By sampling according to this distribution, they achieved a way of sampling with correspondence to the model geometry, which has shown empirical success in achieving high-fidelity generation \cite{chadebec2022geometricperspectivevariationalautoencoders}. The arguments we will present will be in the context of Fisher information statistical theory. Our experiments, established in an alternative view, will not attempt to draw from a similar distribution for FID reasons as they do in this work.

\vspace{2mm}

\noindent An important caveat with this work is here the Riemannian metric has the same intrinsic dimension $d \times d$ as the latent space $d$, thus $G$ does not characterize low-dimensional geometry, but rather geometry of the same space in which the manifold exists. In particular, a manifold exists in the latent space, but it is not low-dimensional, but rather describes the latent space curvature.

\vspace{2mm}

\noindent Allow us to discuss the Fisher information metric in simple, baseline cases, which we will work with. The Fisher information metric is defined, for decoder geometry, via \cite{arvanitidis2022pullinginformationgeometry}
\begin{align}
g_{\alpha \beta} & = \EX_{x \sim p(x|z)} [ \frac{\partial}{\partial z^{\alpha}} \log p(x | z)\frac{\partial}{\partial z^{\beta}}(\log p(x|z))^T ] 
\\
& =  \int_{\mathcal{X}} \Bigg[ (\nabla_z \log p(x | z) )(\nabla_z \log p(x|z))^T \Bigg]_{\alpha \beta} p(x | z) dx  .
\end{align}
For Gaussian likelihood $p(x|z) = \mathcal{N}(x; \mu(z), \Sigma)$, the log-likelihood, assuming constant variance, is
\begin{align}
\log p_{\theta}(x|z) = -\frac{1}{2} (x - \mu(z) )^T \Sigma^{-1} ( x - \mu(z)) + \text{constants} .
\end{align}
Notice
\begin{align}
\partial_{z_i} \log p_{\theta}(x|z) \propto (x - \mu(z))^T \Sigma^{-1} \partial_{z_i} \mu(z) .
\end{align}
Thus the Fisher information metric is
\begin{align}
g_{ij} = \EX [ (\partial_{z_i} \log p (x|z)(\partial_{z_j} \log p(x|z) )] & = \partial_{z_i} \mu^T(z) \Sigma^{-1} \underbrace{ \EX [ (x - \mu)(x- \mu)^T] }_{= \Sigma} \Sigma^{-1} \partial_{z_j} \mu(z) 
\\
& = \partial_{z_i} \mu^T(z) \Sigma^{-1} \partial_{z_j} \mu(z).
\end{align}
The partial derivatives run over the elements of $z$. Equivalently,
\begin{align}
\label{eqn:fisher_metric}
g(z) = J_{\mu}^T(z) \Sigma^{-1} J_{\mu}(z) .
\end{align}

\section{Our work: complex decoder geometry}

Let $K : M \rightarrow \mathbb{R}$ be a Kähler potential, where $M$ is a complex manifold. Let $h$ be a Hermitian metric $h$ such that
\begin{align}
h_{\alpha \overline{\beta}}(z) = \partial_{\alpha} \partial_{\overline{\beta}} K(z,\overline{z}), \ \ \ \ \text{with associated Kähler form} \ \ \ \ \ \omega = i \partial \overline{\partial} K ,
\end{align}
where $(z,\overline{z})$ is within a local coordinate chart and
\begin{align}
K = K(z^1(p),\hdots,z^m(p), \overline{z}^1(p),\hdots,\overline{z}^m(p)) .
\end{align}
The Kähler form is such that
\begin{align}
\omega = i h_{\alpha \overline{\beta}} dz^{\alpha} \wedge d\overline{z}^{\beta}
\end{align}
and the Hermitian metric decomposes into real and imaginary parts \cite{Rowland2025KaehlerForm}
\begin{align}
h = g - i \omega
\end{align}
is the Hermitian metric for Riemannian metric $g$. Note that there are various conventions, so we choose the above.

\subsection{The Fisher information metric in the complex case}

It is known from statistical information theory that the intrinsic decoder geometry is rooted in the Fisher information metric \cite{cho2023hyperbolicvaelatentgaussian} \cite{arvanitidis2022pullinginformationgeometry} \cite{Zacherl_2021} \cite{chadebec2022geometricperspectivevariationalautoencoders}. In this section, we will derive the Fisher information metric that appears in the VAE but with a complex Gaussian distribution assumption. We will use the following definition in our proof: the probability density of a complex normal random variable is, for $P = \overline{\Sigma} - C^{\dagger} \Sigma^{-1} C$,
\begin{align}
f(z)
&= \frac{1}{\pi^{n} \sqrt{ \det(\Sigma)\det(P) }} 
\exp\!\left\{
-\frac{1}{2}
\begin{bmatrix}
z - \mu \\
\overline{z} - \overline{\mu}
\end{bmatrix}^{{\dagger}}
\begin{bmatrix}
\Sigma & C \\
\overline{C} & \overline{\Sigma}
\end{bmatrix}^{\!-1}
\begin{bmatrix}
z - \mu \\
\overline{z} - \overline{\mu}
\end{bmatrix}
\right\} .
\end{align}
We will take $C=0$, which is standard in a VAE, thus the factor of $1/2$ in the exponent is removed. We will always take $\Sigma$ to be Hermitian and diagonal, i.e.
\begin{align}
\label{eqn:covariance_properties}
\Sigma^{\dagger} = \Sigma = \overline{\Sigma}.
\end{align}
The Hermitian quality is a standard fact for complex Gaussians \cite{Hankin2015}, and the complex conjugate quality follows since $\Sigma$ is (typically) diagonal in a VAE.

\vspace{2mm}

\noindent Generally it is the case that a Riemannian metric in a generative model has correspondence to a Fisher information metric. We argue that so does a Hermitian metric under the relation
\begin{align}
\color{RoyalBlue}\boxed{ \color{black}
\rule{0pt}{5mm}\;
\ \ \ \ \ h_{\alpha \overline{\beta}} = \EX \Big[ \partial_{\alpha} \log p \partial_{\overline{\beta}} \log p \Big] .   \ \ \ \ \  \;
\rule[-3mm]{0pt}{0pt}
}
\end{align}
This result follows directly from \cite{gnandi2024kahlermetricfisherinformation}, Theorem 1.4, under the prerogative that this metric is also Kähler, i.e. can be derived from a Kähler potential. From a machine learning perspective, this idea is new. From a physics perspective, it is not. This idea is largely discussed in \cite{Facchi_2010}, which is a foundational reference for complex Fisher metrics. We refer to \cite{Contreras_2021} \cite{Contreras_2016} for other relevant literature.

\vspace{2mm}

\noindent \textbf{Theorem 1.} (a known result; mentioned in \cite{1519691}; the proof is ours) Suppose $x|z \sim \mathcal{CN}(\mu(z),\Sigma(z))$ (in particular, the relation matrix is $C=0$) and that $z,\overline{z}$ are independent. Then the Fisher information Hermitian metric on the decoded latent space, up to constants, is 
\begin{align}
\label{eqn:fisher_complex_metric}
h_{\alpha \overline{\beta}}(z) = 2 \text{Re} (\partial_{\alpha} \mu)^{\dagger} \Sigma^{-1} (\partial_{\overline{\beta}} \mu) +  \text{Tr}( \Sigma^{-1}(\partial_{\alpha} \Sigma) \Sigma^{-1} (\partial_{\overline{\beta}} \Sigma) ) .
\end{align}

\vspace{2mm}

\noindent \textit{Remark.} The Fisher information metric for a Euclidean Gaussian $X \sim \mathcal{N}(x;,\mu(x),\Sigma(x))$ is highly well known \cite{casella2002statistical}, but not studied as often in the complex case. A key difference in the complex case is the fact of $1/2$ appearing in the real-valued case, whereas here it does not due to the complex multivariate Gaussian density definition. 

\vspace{2mm}

\noindent \textit{Proof.} The log-likelihood with trivial relation matrix is
\begin{align}
\log p(x|z) = - (x - \mu)^{\dagger} \Sigma^{-1} (x - \mu)  -  \log \text{det} \Sigma .
\end{align}
Let $\partial_{\gamma}$ denote the element-wise derivative, i.e.
\begin{align}
\partial_{\gamma} \mu = \frac{\partial \mu}{\partial z^{\gamma}} = ( \frac{ \partial \mu_1 }{ \partial z^{\gamma} } , \hdots, \frac{\partial \mu_m }{ \partial z^{\gamma} } ).
\end{align}
Note the following derivative identities:
\begin{align}
& \partial_{\alpha} \Sigma^{-1} = - \Sigma^{-1} (\partial_{\alpha } \Sigma) \Sigma^{-1}
\\
& \partial_{\alpha }\log \text{det} \Sigma = \text{Tr}(\Sigma^{-1} \partial_{\alpha } \Sigma ) .
\end{align}
Now,
\begin{align}
\partial_{\alpha} \log p & = (\partial_{\alpha} \mu)^{\dagger} \Sigma^{-1} (x - \mu) + (x - \mu)^{\dagger} \Sigma^{-1} \partial_{\alpha} \mu + (x-\mu)^{\dagger} \underbrace{  \Sigma^{-1} ( \partial_{\alpha} \Sigma) \Sigma^{-1} }_{=A} ( x - \mu ) - \underbrace{  \text{Tr}(\Sigma^{-1} \partial_{\alpha} \Sigma ) }_{=t_{\alpha}}
\\
\partial_{\overline{\beta}} \log p  & = 
 (\partial_{\overline{\beta}} \mu)^{\dagger} \Sigma^{-1} (x - \mu) + (x - \mu)^{\dagger} \Sigma^{-1} (\partial_{\overline{\beta}} \mu)   + (x-\mu)^{\dagger} \underbrace{  \Sigma^{-1} ( \partial_{\overline{\beta}} \Sigma) \Sigma^{-1} }_{=B} ( x - \mu )   - \underbrace{  \text{Tr}(\Sigma^{-1} \partial_{\overline{\beta}} \Sigma ) }_{=t_{\overline{\beta}}} .
\end{align}
Set $v = x - \mu$. Observe the following \cite{petersen2012matrix}:
\begin{align}
& \EX[v] = 0
\\
& \EX[ v v^{\dagger}] = \Sigma
\\
& \EX[ v^{\dagger} M v] = \text{Tr}(M\Sigma)
\\
& \EX[ v^{\dagger} M v  v^{\dagger} N v] = \text{Tr}(M\Sigma N \Sigma) + \text{Tr}(M\Sigma) \text{Tr}(N \Sigma)
\\
& \EX [ a^{\dagger} v v^{\dagger} M v v^{\dagger} b ] = a^{\dagger} \Sigma M \Sigma b + (a^{\dagger} \Sigma b)\text{Tr}(M\Sigma) 
\\
& \EX[ a^{\dagger} M v v^{\dagger} N b] = a^{\dagger} M \Sigma N b .
\end{align}
Now, it is known from the Fisher information
\begin{align}
h_{\alpha \overline{\beta}} & = \EX[ \partial_{\alpha} \log p \partial_{\overline{\beta}} \log p ]
\\[2em]
& = \EX \Bigg[ (\partial_{\alpha} \mu)^{\dagger} \Sigma^{-1} v (\partial_{\overline{\beta}} \mu)^{\dagger} \Sigma^{-1} v +  (\partial_{\alpha} \mu)^{\dagger} \Sigma^{-1} v v^{\dagger} \Sigma^{-1}(\partial_{\overline{\beta}}  \mu) + (\partial_{\alpha} \mu)^{\dagger} \Sigma^{-1} v v^{\dagger} B v - (\partial_{\alpha} \mu)^{\dagger} \Sigma^{-1} v t_{\overline{\beta}}
\\
& \ \ \ \ \ \ + v^{\dagger} \Sigma^{-1} (\partial_{\alpha} \mu) (\partial_{\overline{\beta}}  \mu)^{\dagger} \Sigma^{-1} v +  v^{\dagger} \Sigma^{-1} (\partial_{\alpha}  \mu) v^{\dagger} \Sigma^{-1} (\partial_{\overline{\beta}}  \mu)  + v^{\dagger} \Sigma^{-1} (\partial_{\alpha} \mu) v^{\dagger} B v -  v^{\dagger} \Sigma^{-1} (\partial_{\alpha} \mu) t_{\overline{\beta}} 
\\
& \ \ \ \ \ \ + v^{\dagger} A v (\partial_{\overline{\beta}} \mu)^{\dagger} \Sigma^{-1} v + v^{\dagger} A v v^{\dagger} \Sigma^{-1} (\partial_{\overline{\beta}}  \mu) + v^{\dagger} A v v^{\dagger} B v - v^{\dagger} A v t_{\overline{\beta}} 
\\
& \ \ \ \ \ \ - t_{\alpha} (\partial_{\overline{\beta}}  \mu)^{\dagger} \Sigma^{-1} v - t_{\alpha} v^{\dagger} \Sigma^{-1} (\partial_{\overline{\beta}}  \mu) - t_{\alpha} v^{\dagger} B v + t_{\alpha} t_{\overline{\beta}}  \Bigg]
\\[2em]
& = 0 + (\partial_{\alpha} \mu)^{\dagger} \Sigma^{-1} (\partial_{\overline{\beta}} \mu) + 0 + 0  + \text{Tr}(\Sigma^{-1} (\partial_{\alpha}\mu)(\partial_{\overline{\beta}} \mu)^{\dagger} ) + 0 + 0 + 0 + 0 + 0\\
& \ \ \ \ \ \ + \text{Tr}(A \Sigma B \Sigma) + \text{Tr}(A \Sigma) \text{Tr}(B \Sigma) - t_{\overline{\beta}} \text{Tr}(A\Sigma) - 0 + 0 - t_{\alpha} \text{Tr}(B\Sigma) + t_{\alpha} t_{\overline{\beta}}
\\[2em]
& =  (\partial_{\alpha} \mu)^{\dagger} \Sigma^{-1} (\partial_{\overline{\beta}} \mu) +\text{Tr}(\Sigma^{-1} (\partial_{\alpha}\mu)(\partial_{\overline{\beta}} \mu)^{\dagger} ) + \text{Tr}(A \Sigma B \Sigma) 
\\[2em]
& = 2 \text{Re} (\partial_{\alpha} \mu)^{\dagger} \Sigma^{-1} (\partial_{\overline{\beta}} \mu) + \text{Tr}( \Sigma^{-1} \partial_{\alpha } \Sigma \Sigma^{-1} \partial_{\overline{\beta}} \Sigma) .
\end{align}
This completes the proof.

\noindent $ \square $

\vspace{2mm}

\noindent In practice, the metric of equation \ref{eqn:fisher_complex_metric} is nontrivial to sample from since it requires differentiation of $\Sigma$. We will attempt to construct a metric that has greater ability to be sampled in the following section.

\subsection{The Fisher information metric and the Hessian of the KL divergence as a Kähler potential}

It is well known in statistical theory that the Fisher information is the Hessian of the Kullback-Leibler (KL) divergence. Let the Fisher information with respect to the paramater be defined as
\begin{align}
I = \EX \Big[ (\nabla_{\theta} \log p_{\theta}(x))(\nabla_{\theta} \log p_{\theta}(x))^T \Big] .
\end{align}
As Riemann integral and Lebesgue integrals respectively, the KL divergence is
\begin{align}
\text{KL} ( p_{\theta}(x) \parallel p_{\theta'}(x) ) = \int p_{\theta}(x) \log \frac{p_{\theta}(x)}{p_{\theta'}(x) } dx  = \int \log \frac{dP_{\theta}}{dP_{\theta'}} dP_{\theta} .
\end{align}
In particular, it is known \cite{Zuo2017}
\begin{align}
I = \text{Hess}_{\theta'} \text{KL}(p_{\theta}(x) \parallel p_{\theta'}(x) ) \Bigg|_{\theta' = \theta} .
\end{align}
However, a Kähler potential necessitates
\begin{align}
h_{\alpha \overline{\beta}} = \partial_{\alpha} \partial_{\overline{\beta}} K(z,\overline{z}) ,
\end{align}
i.e. the partial derivative is with respect to $z$ coordinate, whereas the Fisher information derivatives are taken with respect to the neural network parameters in the above. Thus, if we take $z$ as the parameter in the above, we have
\begin{align}
h_{\alpha \overline{\beta}} = \text{Hess}_{z', \alpha, \overline{\beta}} \text{KL}(p(x|z) \parallel p(x|z')) \Bigg|_{z' = z} =  \partial_{\alpha} \partial_{\overline{\beta}} \text{KL}(p(x|z) \parallel p(x|z') ) \Bigg|_{z' = z}  = \partial_{\alpha } \partial_{\overline{\beta}} K(z,\overline{z})  ,
\end{align}
and so we take our Kähler potential as
\begin{align}
\color{RoyalBlue}\boxed{ \color{black}
\rule{0pt}{5mm}\;
\ \ \ \ \ K(z,\overline{z}) =\text{KL}(p(x|z) \parallel p(x|z') )   \ \ \ \ \  \;
\rule[-3mm]{0pt}{0pt}
}
\end{align}
subsequently evaluated at $z'=z$ once the (complex) Hessian is taken.

\vspace{2mm}

\noindent Moreover, $p(x|z) \sim \mathcal{N}_{\mathbb{C}}(x; \mu(z), \Sigma(z))$ is complex Gaussian, and the KL divergence between complex Gaussians has a closed form. It is known
\begin{align}
\text{KL}\big(p(x|z) \parallel p(x|z')\big)
= 
\mathrm{Tr}\!\big(\Sigma(z')^{-1} \Sigma(z)\big)
+ (\mu(z') - \mu(z))^{\dagger} \Sigma(z')^{-1} (\mu(z') - \mu(z))
- n
+ \ln\!\frac{|\Sigma(z')|}{|\Sigma(z)|}
.
\end{align}
When the covariance is diagonal, we get
\begin{align}
\text{KL}\big(p(x|z) \parallel p(x|z')\big)
=  \sum_{k=1}^n
\Bigg[
\frac{\sigma_k^2(z)}{\sigma_k^2(z')}
+ \frac{|\mu_k(z) - \mu_k(z')|^2}{\sigma_k^2(z')}
- 1
+ \ln\!\frac{\sigma_k^2(z')}{\sigma_k^2(z)}
\Bigg].
\end{align}
Thus, by matching,
\begin{align}
\partial_{\alpha} \partial_{\overline{\beta}} \ \text{KL}\big(p(x|z) \parallel p(x|z')\big)
\Bigg|_{z'=z}
= 2 \text{Re} (\partial_{\alpha} \mu)^{\dagger} \Sigma^{-1} (\partial_{\overline{\beta}} \mu) +  \text{Tr}( \Sigma^{-1}(\partial_{\alpha} \Sigma) \Sigma^{-1} (\partial_{\overline{\beta}} \Sigma) ) ,
\end{align}
or equivalently in line element form
\begin{align}
ds^2 = 2 \text{Re} \ d\mu^{\dagger} \Sigma^{-1} d\mu +  \text{Tr}(\Sigma^{-1} d\Sigma \Sigma^{-1} d\Sigma ) .
\end{align}
We refer to \cite{amari2000methods} \cite{mandolesi2025hermitiangeometrycomplexmultivectors} for additional relevant literature.

\subsection{A log-likelihood sum potential induces a metric dependent on the Fisher information metric in the continuum limit}
\label{sec:proposed_metric}

\begin{figure}[htbp]
  \centering
  \vspace{0mm}
  \includegraphics[scale=0.55]{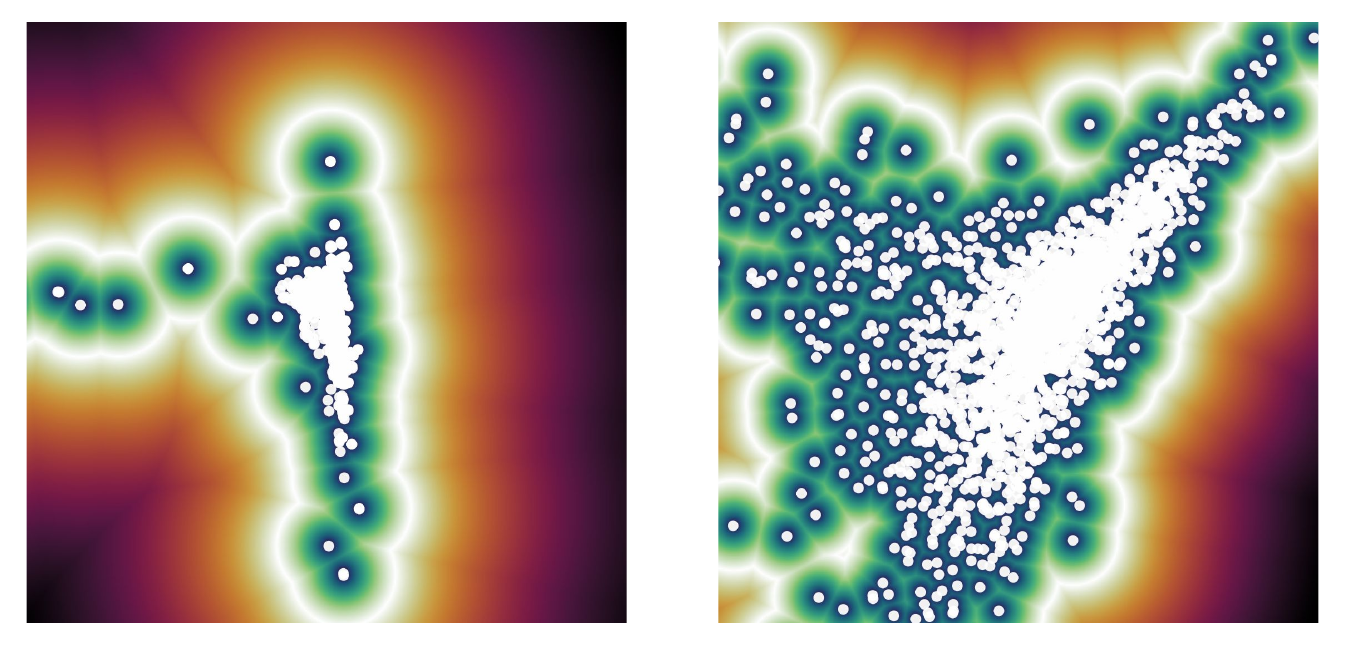}
  \caption{We illustrate a component reduction method in order to examine the effect of curvature on the sampling procedure on (left) a complex Gaussian-sampled latent space; (right) a metric-sampled latent space according to our method. We construct a Laplacian Eigenmaps embedding on a curvature-weighted graph based on the Eigenproblem $Lv = \lambda Dv $ (note that a Gaussian solves $\lambda = 0, v=1$). In particular, we plot $(v_1(i),v_2(i))$ coordinate eigenfunctions that minimize the curvature-weighted energy problem subject to eigenconstraints. A Gaussian space is solved nearly trivially (Gaussian plots looks like a line, as in left, since one eigenfunction is constant), and curvature is conveyed (via a curved manifold) as in the right. As we can see, there is notable difference.}
  \label{fig:latent_curvature}
\end{figure}

The primary goal of this section is that we will show there exists a Kähler potential that induces a Hermitian metric that depends on the true Fisher information metric. In particular, our method is more justified theoretically rather than simply "replacing" terms in \ref{eqn:fisher_complex_metric}, since if we swap terms for things more computable, it will not necessarily hold that Kähler geometry exists. Thus, this section will attempt to justify we can replace the Hermitian metric of \ref{eqn:fisher_complex_metric} with something that is valid as a metric that is simultaneously more computable under assumptions.

\vspace{2mm}

\noindent We provide some motivation for this section. We begin by sampling a collection in the latent space. We refer to this as the previous samples. In this section, we anticipate a sampled point $z$ is sufficiently "close" to a previously sampled mean, i.e. the previous sample is sufficiently dense. Under this assumption, we operate so that covariance is estimated sufficiently well under this density. Moreover, the metric we provide will be an expected value analog in the previous section, more or less. We will argue, under this density of points and due to exponential convergence properties, the weights of our expected value will be arranged in such a way that the weight of the "distant" points to the newly sampled point is significantly larger than those of previously sampled points far away. Thus our expectation will coincide with the derivation in Appendix \ref{app:appendix_a}, and has dependence on the Fisher metric. Moreover, we still maintain validity of the metric. 

\vspace{2mm}

\noindent We provide some additional motivation. We will use Gaussian-like functions, but with positive quadratic form instead of the prototypical negative. Thus, distant points will be rewarded with high function value, whereas the global maximum of a Gaussian (or Gaussian-like) is (typically) at its mean. This is atypical so we motivate this for the following reasons. Our metric is valid under this positive quadratic form, whereas adding negative signs to the Kähler potential is not necessarily so. Moreover, our methods here match the methods as in Appendix \ref{app:appendix_a}, thus the laws of total expectation and covariance hold between the two derivations. Were were to add negative signs to our Kähler potential, this is not the case. Thus, our methods serve more as replacements/proxies to the Fisher information metric rather than approximations. The edge posterior points will be rewarded in sampling among posterior variational sampling. In practice, we find this complements the loss of \ref{eqn:loss_metric} well, and our computational results are nontrivially meaningful as in Table \ref{tab:method_results_table}.

\vspace{2mm}

\noindent In our experiments, we will add to the loss
\begin{align}
\mathcal{L} = \EX[ z^{\dagger} z - \log \text{det} (h) ] .
\end{align}
This will encourage a larger metric as $z$ is distant from the origin. This contributes to justification of our methods.

\vspace{2mm}

\noindent Let $z$ be a latent point, $x$ the decoder, and $\mu$ be the map of a mean of a Gaussian of the map in the latent space. Let us take Gaussian weights
\begin{align}
\Omega_i(z) = \exp \Bigg\{ \frac{1}{\rho^2} (x(z)-\mu_i)^{\dagger} \Sigma_i^{-1} (x(z) - \mu_i ) \Bigg\}, \ \ \ \ \ \Psi_i = (x(z)-\mu_i)^{\dagger} \Sigma_i^{-1} (x(z) - \mu_i ) ,
\end{align}
with Kähler potential, which is a log-likelihood of Gaussian components,
\begin{align}
\label{eqn:Kähler_potential}
K(z,\overline{z}) = \rho^2 \log \Big( \sum_i a_i \Omega_i(z)  \Big) , \ \ \ \ \ \sum_i a_i = 1, a_i > 0, 
\end{align}
where we will often take $a_i = \frac{1}{N}$. We refer to \cite{korotkine2024hessiangaussianmixturelikelihoods} for relevant statistical literature on this potential. This Kähler potential is distinct from statistical literature. Taking a logarithm of a mixture sum is not frequent in literature. From a statistical perspective, it poses computational issues \cite{rudin2015gmmem}. For instance, the sum is in the interior of the logarithm and makes statistical computational difficult, if even relevant. In literature, a similar Kähler potential is studied in \cite{xie2023complexrecurrentvariationalautoencoder} from a geometry perspective, but without the Gaussian component. We emphasize the positive sign in the above quadratic form, as this matches the theoretical derivation involving $\Psi$ as in Appendix \ref{app:appendix_a} with
\begin{align}
\Psi \ (\text{here}) = \Psi \ (\text{appendix}) - \text{other terms} .
\end{align}
The derivation with $-\Psi$ will not yield the right final signs.

\vspace{2mm}

\noindent Our Kähler metric is given by
\begin{align}
h_{\alpha \overline{\beta}}(z) = \partial_{\alpha} \partial_{\overline{\beta}} K(z,\overline{z}) .
\end{align}
Let us derive this. Define $v_i := x - \mu_i$. First, notice, in Einstein notation and since $x$ is pluriharmonic,
\begin{gather}
\Psi_i = (\Sigma_i^{-1})_{\alpha \overline{\beta}} v_i^{\alpha} v_i^{\overline{\beta}}
\\
\partial_{\alpha} \Psi_i = (\partial_{\alpha} v_i^{\delta}) (\Sigma_i^{-1})_{\delta \overline{\beta}} v_i^{\overline{\beta}} + v_i^{\delta} (\Sigma_i^{-1})_{\delta \overline{\beta}} (\partial_{\alpha} v_i^{\overline{\beta}} ), \ \ \ \ \ \partial_{\overline{\beta}} \Psi_i = (\partial_{\overline{\beta}} v_i^{\overline{\delta}}) (\Sigma_i^{-1})_{\alpha \overline{\delta}} v_i^{\alpha} +  v_i^{\overline{\delta}} (\Sigma_i^{-1})_{\alpha \overline{\delta}} ( \partial_{\overline{\beta}} v_i^{\alpha} )
\\
\partial_{\alpha} \partial_{\overline{\beta}} \Psi_i = (\partial_{\alpha} v_i^{\delta})(\Sigma_i^{-1})_{\delta \overline{\gamma}}(\partial_{\overline{\beta}} v_i^{\overline{\gamma}} ) + (\partial_{\overline{\beta}} v_i^{\delta})(\Sigma_i^{-1})_{\delta \overline{\gamma}}(\partial_{\alpha} v_i^{\overline{\gamma}} ) .
\end{gather}
Now, using the exponential (Wirtinger) derivative and the product rule,
\begin{align}
\partial_{\overline{\beta}} \Omega_i & = \partial_{\overline{\beta}} \Big( e^{\Psi_i / \rho^2 } \Big) = \frac{1}{\rho^2} ( \partial_{\overline{\beta}} \Psi_i) \Omega_i 
\\
\partial_{\alpha} \partial_{\overline{\beta}} \Omega_i  & = \frac{1}{\rho^2} \Big[ \partial_{\alpha} \Omega_i \partial_{\overline{\beta}} \Psi_i  + \Omega_i \partial_{\alpha} \partial_{\overline{\beta}} \Psi_i \Big] = \frac{\Omega_i}{\rho^2} \Big[  \frac{1}{\rho^2}  \partial_{\alpha} \Psi_i \partial_{\overline{\beta}} \Psi_i  + \partial_{\alpha} \partial_{\overline{\beta}} \Psi_i \Big] 
\\
& = \frac{\Omega_i}{\rho^2} \Big[  \frac{1}{\rho^2} [ (\partial_{\alpha} v_i^{\delta}) (\Sigma_i^{-1})_{\delta \overline{\gamma}} v_i^{\overline{\gamma}} + v_i^{\delta} (\Sigma_i^{-1})_{\delta \overline{\gamma}} (\partial_{\alpha} v_i^{\overline{\gamma}} ) ][ (\partial_{\overline{\beta}} v_i^{\overline{\delta}}) (\Sigma_i^{-1})_{\gamma \overline{\delta}} v_i^{\gamma}
\\
& \ \ \ \ \ \ \ \ \ +  v_i^{\overline{\delta}} (\Sigma_i^{-1})_{\gamma \overline{\delta}} ( \partial_{\overline{\beta}} v_i^{\gamma} ) ] + (\partial_{\alpha} v_i^{\delta})(\Sigma_i^{-1})_{\delta \overline{\gamma}}(\partial_{\overline{\beta}} v_i^{\overline{\gamma}} ) + (\partial_{\overline{\beta}} v_i^{\delta})(\Sigma_i^{-1})_{\delta \overline{\gamma}}(\partial_{\alpha} v_i^{\overline{\gamma}} )  \Big] := \Xi_i .
\end{align}
Thus, we will endow $z$ to be on a compact subset of $\mathbb{C}^d$, then by uniform convergence by the Weierstrass M-test, we can exchange differentiation and the infinite sum
\begin{align}
h_{\alpha \overline{\beta}} & = \partial_{\alpha} \partial_{\overline{\beta}} \Big(  \rho^2 \log \Big(  \sum_i a_i \Omega_i(z)  \Big) \Big)
\\
& = \rho^2 \partial_{\alpha} \Big(  \frac{1}{\sum_i a_i \Omega_i(z) } \sum_i a_i \partial_{\overline{\beta}} \Omega_i(z) \Big) 
\\
& = \rho^2  \Bigg(  -\frac{ \sum_i a_i  \partial_{\alpha} \Omega_i(z)}{(\sum_i a_i \Omega_i(z))^2 } \sum_i a_i \partial_{\overline{\beta}} \Omega_i(z) +  \frac{1}{\sum_i a_i \Omega_i(z) } \sum_i a_i \partial_{\alpha} \partial_{\overline{\beta}} \Omega_i(z)  \Bigg) 
\\
& = \rho^2  \Bigg(  \frac{ -1}{(\sum_i a_i \Omega_i(z))^2 }  (\sum_i a_i \frac{1}{\rho^2}\Omega_i \partial_{\overline{\beta}} \Psi_i) (\sum_i a_i \frac{1}{\rho^2}\Omega_i \partial_{\alpha} \Psi_i)  +  \frac{1}{\sum_i a_i \Omega_i(z) } \sum_i a_i \Xi_i  \Bigg)  .
\end{align}
This is the Kähler metric of the complex VAE. Let us justify this more. Now, notice the following limits, which we will justify shortly:
\begin{align}
\frac{\sum_i a_i \Omega_i(z) \partial_{\alpha} \Psi_i}{\sum_i a_i \Omega_i(z)} & \stackrel{N \rightarrow \infty}{\longrightarrow}
\EX_{w} [ \partial_{\alpha} \Psi]
\\
\frac{\sum_i a_i \Omega_i(z) \partial_{\overline{\beta}} \Psi_i}{\sum_i a_i \Omega_i(z)} & \stackrel{N \rightarrow \infty}{\longrightarrow} \EX_{w} [ \partial_{\overline{\beta}} \Psi]
\\
\frac{1}{\sum_i a_i \Omega_i(z)} \sum_i a_i \Xi_i & \stackrel{N \rightarrow \infty}{\longrightarrow} \frac{1}{\rho^2} \EX_{w} [ \frac{1}{\rho^2} \partial_{\alpha} \Psi \partial_{\overline{\beta}} \Psi + \partial_{\alpha} \partial_{\overline{\beta}} \Psi] .
\end{align}
Thus,
\begin{align}
& h_{\alpha \overline{\beta}}  =  \rho^2 \Bigg( \frac{-1}{\rho^4} \EX_{w}[\partial_{\alpha}\Psi] \EX_{w}[\partial_{\overline{\beta}}\Psi] + \frac{1}{\rho^2} \EX_{w} [ \frac{1}{\rho^2} \partial_{\alpha} \Psi \partial_{\overline{\beta}} \Psi + \partial_{\alpha} \partial_{\overline{\beta}} \Psi] \Bigg)
\\
& \ \ \ \ \ = \EX_{w} [ \partial_{\alpha} \partial_{\overline{\beta}} \Psi] + \frac{1}{\rho^2} \Big( \EX_{w} [ \partial_{\alpha} \Psi \partial_{\overline{\beta}} \Psi ] - \EX_{w}[\partial_{\alpha}\Psi] \EX_{w}[\partial_{\overline{\beta}}\Psi] \Big)
\\
\label{eqn:Kähler_metric}
& \color{RoyalBlue}\boxed{ \color{black}
h_{\alpha \overline{\beta}} = \EX_{w} [ \partial_{\alpha} \partial_{\overline{\beta}} \Psi] + \frac{1}{\rho^2} \text{Cov}(\partial_{\alpha} \Psi, \partial_{\overline{\beta}} \Psi) }
\\
& = \sum_i w_i(z) \partial_{\alpha} \partial_{\overline{\beta}} \Psi_i(z) + \frac{1}{\rho^2} \Bigg(  \sum_i w_i (\partial_{\alpha} \Psi_i(z) \partial_{\overline{\beta}} \Psi_i(z) ) - \Big[ \sum_i w_i \partial_{\alpha} \Psi_i(z) \Big] \Big[ \sum_i w_i \partial_{\overline{\beta}} \Psi_i(z) \Big]  \Bigg) .
\end{align}
We have
\begin{align}
w_i(z) \stackrel{!}{=} \frac{a_i e^{\Psi_i(z) / \rho^2}}{ \sum_j a_j e^{\Psi_j(z) / \rho^2}}, \ \ \ \ \ \EX_{w} [ X ] = \sum_i w_i(z) X(z) .
\end{align}
It is known that $\text{Hess} = J^{\dagger} J$ is reasonable. Thus, we can take
\begin{align}
h_{\alpha \overline{\beta}} = (1 + \frac{1}{\rho^2} ) \EX_{w} [ \Phi ] -  \frac{1}{\rho^2} \EX_{w}[\partial_{\alpha}\Psi] \EX_{w}[\partial_{\overline{\beta}}\Psi] 
\end{align}
if desired for suitable Hessian surrogate $\Phi$.

\vspace{2mm}

\noindent It is crucial to understand our derivation for the potential has used $+\Psi$. For this reason, we will always work with compact sets around the origin, which is typical in a VAE due to the KL regularization. Note that this is potentially numerically unstable. Thus, an alternative solution is to take
\begin{align}
\label{eqn:improved_weights}
w_i(z) = \frac{a_i e^{(\Psi_i(z) - M)/ \rho^2}}{ \sum_j a_j e^{(\Psi_j(z) - M) / \rho^2}}
\end{align}
for suitable $M$, i.e. $M = \max_i \Psi_i$ for stability reasons.

\vspace{2mm}

\noindent Note that we reconcile the proof as in here and Appendix \ref{app:appendix_a} because
\begin{align}
\label{eqn:iterated_expectation}
\tilde{h}_{\alpha \overline{\beta}}(z) \propto \mathbb{E}_{w(i)} \left[ \mathbb{E}_{x|z,i}[\partial_{\alpha} \partial_{\overline{\beta}} \Psi_i] + \frac{1}{\rho^2} \mathrm{Cov}_{x|z,i}(\partial_{\alpha} \Psi_i, \partial_{\overline{\beta}} \Psi_i) \right] + \frac{1}{\rho^2} \mathrm{Cov}_{w(i)} \left( \mathbb{E}_{x|z,i} [ \partial_{\alpha} \Psi_i ], \mathbb{E}_{x|z,i} [ \partial_{\overline{\beta}} \Psi_i ] \right) ,
\end{align}
where $\tilde{h}$ is a true target metric, which is an asymptotic identity, thus our methods serves as approximations. We refer to Appendix \ref{sec:exp_and_cov_relations} for full details.

\vspace{2mm}

\noindent The argument for our expectations come from weighted ergodic theory. Consider a sample $\{\Gamma_i(z)\}_{i=1}^{\infty}$ with weights $\EX [ w] < \infty, \EX [ w\Gamma] < \infty$. Taking the limit, we have
\begin{align}
\lim_{N \rightarrow \infty} \frac{ \sum_{i=1}^N w_i \Gamma_i(z) }{ \sum_{i=1}^N w_i} {\longrightarrow} \EX_{\mu} [ \Gamma(z)] = \int_{\Omega} \Gamma(z,\theta) d\mu(\theta) .
\end{align}
We borrow from Theorem 11 in \cite{Yoshimoto1974}. Since our domain is compact, $\Gamma_i(z) = \Psi_i(z)$ is $L^1$ and take $h$ as in \cite{Yoshimoto1974} to be $h(z) =1$, which is certainly  measurable for suitable probability measure on $z$. For example, take an open ball enclosing $1$, which is in the Borel $\sigma$-algebra, thus the preimage is the entire space of $\mathbb{R}^d$, which is in the $\sigma$-algebra of the domain. As in \cite{Yoshimoto1974}, take $\alpha_i = w_i$. By the result of the Theorem, this limit exists and is finite on the numerator's support $\{ z : \sum_i w_i \Gamma_i(z) \}$. Since $\Gamma_i$ is a quadratic form for us, it is nonnegative ($\Sigma$ is positive semi-definite).

\vspace{2mm}

\noindent Let us also justify more our infinite series converges so that the Weierstrass M-test applies. In particular, let us examine
\begin{align}
\rho^2  \lim_{N \rightarrow \infty} \log \Big(  \sum_{i=1}^N a_i \Omega_i(z)  \Big) = \rho^2  \log \Big(  \lim_{N \rightarrow \infty} \sum_{i=1}^N a_i \Omega_i(z)  \Big)
\end{align}
on a compact set. The interchange of the log and limit is justified since $\log$ is a continuous function, and the inner series converges by Dirichlet's test.

\vspace{2mm}

\noindent Our metric can also be computed using
\begin{align}
\rho^2 \partial \overline{\partial} \log \Big( \sum_i a_i \Omega_i(z)  \Big) ,
\end{align}
or alternatively, using a fast approximation,
\begin{align}
\rho^2 \Big( J_{(z,\overline{z})} \log \Big( \sum_i a_i \Omega_i(z)  \Big) \Big)^{\dagger} \Big( J_{(z,\overline{z})} \log \Big( \sum_i a_i \Omega_i(z)  \Big) \Big) .
\end{align}
However, the above is currently an ill-defined computation, since it is a Jacobian of a scalar, hence the result is rank one. Thus, we will work with directional derivatives to ensure nontriviality. We can remedy this by taking $F(z) = \Sigma^{-1/2}(x(z) - m(z)), m(z) = \sum_i w_i \mu_i$, and take directional derivatives $g_k(z) = J_F^{\dagger}(z) v^{(k)}$, and then
\begin{align}
\widehat{h}(z) = \frac{\rho^2}{K} \sum_{i=1}^K g_k(z) g_{k}^{\dagger}(z)
\end{align}
is our high-rank proxy. To reiterate, this lists the true Hessian, an ill-defined Jacobian proxy (since it is scalar), and a Hessian proxy that is high-rank.

\vspace{2mm}

\noindent We make some other remarks. Suppose we wanted to attempt to approximate the Fisher metric of equation \ref{eqn:fisher_complex_metric} by using discrete substitutions $\mu_i, \Sigma_i^{-1}$. This is nontrivial because attempting to find the closest point among a collection requires a nearest neighbor algorithm. This is especially nontrivial in large-scale settings, which is the focus of this work. This is already baked into the expectation calculation via the softmax function, so our proposed metric of \ref{eqn:Kähler_metric}, always a valid metric regardless of sampling, has no need for such a nearest-point search algorithm.

\vspace{2mm}

\noindent Our metric indeed discards the $\partial_{\gamma} \Sigma$ terms. It is of interest to attempt to replace these structures in the calculations with a fast approximation. Our proposed Kähler potential discards $\partial_{\gamma} \Sigma$ in the computation but not geometrically. As we show in Appendix \ref{app:appendix_a}, these derivatives are absorbed into the expectation and covariance terms in the purely analytical case. Our proposed method is a discrete approximation to this. We threw away our $\partial_{\gamma} \Sigma$ terms with different sources of curvature, as now the covariance term acts as a substitute for $\partial_{\gamma} \Sigma$.

\vspace{2mm}

\noindent \textbf{Lemma 1.} Suppose $x(z) \approx \mu_i$. Then
\begin{align}
\EX_{w} [ \partial_{\alpha} \partial_{\overline{\beta}} \Psi ] = \sum_i w_i \partial_{\alpha} \partial_{\overline{\beta}} \Psi_i  \approx (\partial_{\alpha} \mu)^{\dagger} \Sigma_j^{-1} ( \partial_{\overline{\beta}} \mu) + (\partial_{\overline{\beta}} \mu)^{\dagger} \Sigma_j^{-1} (\partial_{\alpha} \mu) 
\end{align}
for some $j$.

\vspace{2mm}

\noindent \textit{Proof.} The first equality follows from definition of the expectation. For a single Gaussian, notice $w_j \approx 1$, $w_{i \neq j} \approx 0$ for $x(z) \approx \mu_i$ due to the softmax-type activation, and by definition
\begin{align}
\partial_{\alpha} \partial_{\overline{\beta}} \Psi_i = (\partial_{\alpha} v_j )^{\dagger} (\Sigma_j^{-1} ) (\partial_{\overline{\beta}} v_j ) + (\partial_{\overline{\beta}} v_j )^{\dagger} (\Sigma_j^{-1} ) (\partial_{\alpha} v_j )  ,   
\end{align}
so the result follows. If Gaussians are sufficiently close to one another, take for suitable collection of indices $I$, $\sum_{i \in I} w_i \approx 1$, $w_{j \notin I} \approx 0$.

\noindent $ \square $

\vspace{2mm}

\noindent \textit{Remark.} Indeed, we verified empirically the index of the largest $\Psi$ corresponds to the index of the largest weight, which implies $j \neq i$ typically. In particular, the "winner takes all" in the softmax corresponds to distant $\Psi$.

\vspace{2mm}

\vspace{0mm}
\begin{figure}
  \vspace{0mm}
  \centering
  \includegraphics[scale=0.65]{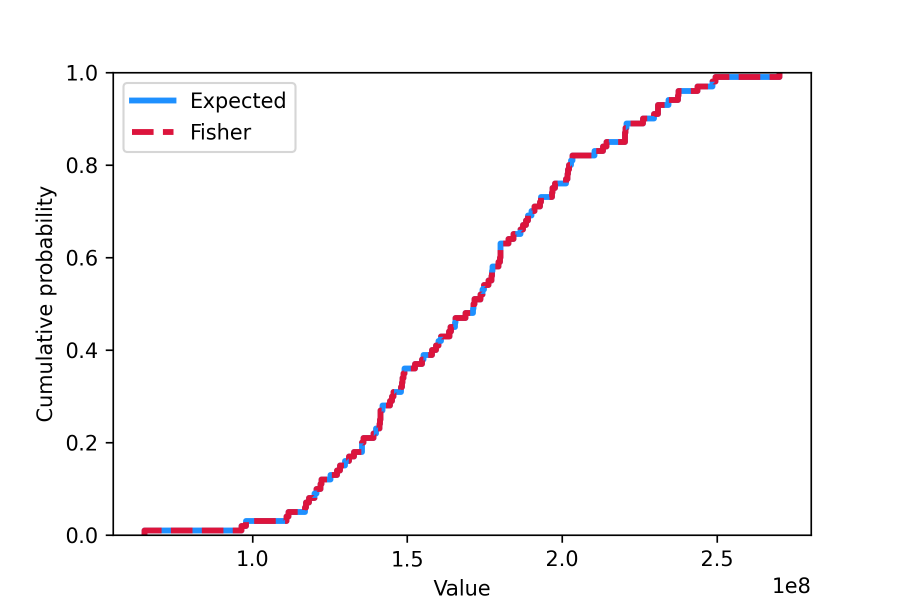}
  \caption{We show distributional equivalence by plotting the cumulative distribution of the $(1,1)$ real element of the first term Fisher information metric $(\partial_{\alpha} \mu)^{\dagger} \Sigma_j^{-1} ( \partial_{\overline{\beta}} \mu) + (\partial_{\overline{\beta}} \mu)^{\dagger} \Sigma_j^{-1} (\partial_{\alpha} \mu) $ along the maximum index versus computing the expectation of $\EX[ \partial_{\alpha} \partial_{\overline{\beta}} \Psi]$ using the softmax. The purpose of this plot is to convey the asymptotic equivalence of Lemma 1, i.e. the softmax will be dominated by a singular weight.}
  \label{fig:dist_equiv}
\end{figure}

\vspace{0mm}
\begin{figure}
  \vspace{0mm}
  \centering
  \includegraphics[scale=0.55]{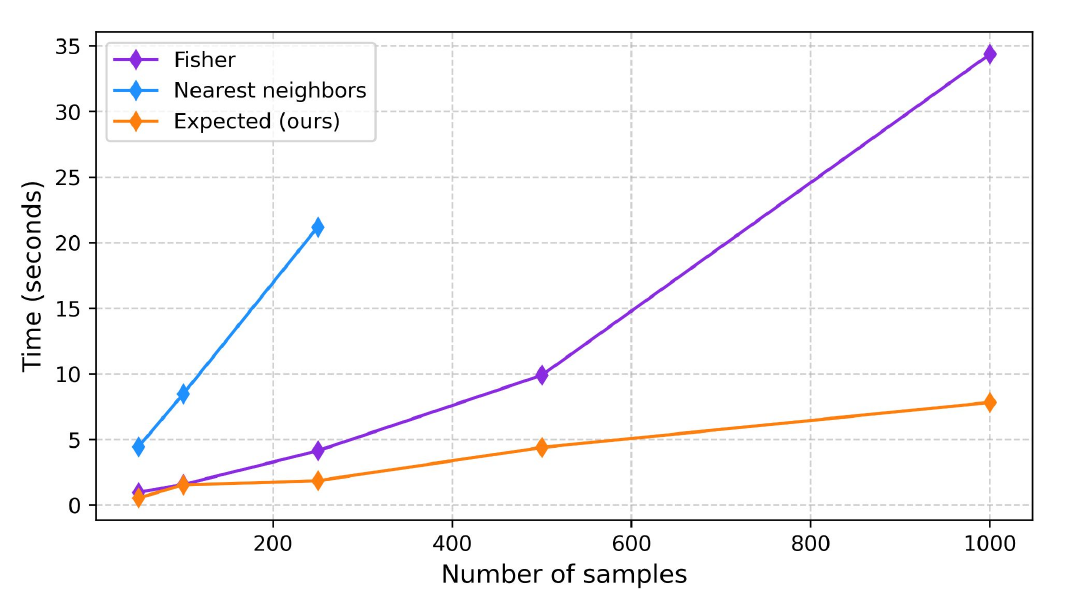}
  \caption{We illustrate runtime in constructing the Hermitian metrics via sampling. We choose $\alpha = \overline{\beta}=1$. We truncate the number of data points for the nearest neighbors search case, since this gave CUDA out of memory errors in high cases. }
  \label{fig:dist_equiv}
\end{figure}

\noindent \textbf{Lemma 2 (a well-known fact).} $h_{\alpha \overline{\beta}}$ is valid as a Hermitian metric and $g$ a Riemannian metric if the potential $\Phi$ is plurisubharmonic. 

\vspace{2mm}

\noindent \textit{Proof.} Since potential $\Phi$ is plurisubharmonic then 
\begin{align}
\partial \overline{\partial} \Phi \succeq 0  \implies \omega = \frac{i}{2} \partial \overline{\partial} \Phi \geq 0 .
\end{align}
Thus, the bilinear form $g(u,v) = \omega(u,Jv)$ is positive semi-definite and a valid Riemannian metric. Moreover, since $h = \partial \overline{\partial} \Phi \succeq 0$, we have $h$ is Hermitian, and so $(g,J,h)$ is valid as a Kähler structure.

\noindent $ \square $

\vspace{2mm}

\noindent \textbf{Theorem 2.} The potential $K$ we defined in \ref{eqn:Kähler_potential} is plurisubharmonic.

\vspace{2mm}

\noindent \textit{Proof.} It is known that that if $\Omega \subseteq \mathbb{C}^d$, and each $u_i$ is plurisubharmonic on $\Omega$, and $\Phi$ is real-valued, increasing, and convex, then $\Phi \circ (u_1,\hdots,u_K)$ is plurisubharmonic. This is found and proven as Proposition 2.1.1 in \cite{Celik2015}. In particular, $u_i$ plurisubharmonic for all $i$ implies is $(u_1,\hdots,u_K)$ is in $\text{PSH}(\Omega)$ in each component. It is known that $\log ( \sum_i e^{x_i} )$ is a convex, and increasing, function over its domain \cite{stackexchange:log-sum-exp-convex}. Thus, it suffices to show $u_i := \Psi_i / \rho^2 + \log a_i $ is plurisubharmonic. Now,
\begin{align}
\partial \overline{\partial} \Psi_i = J_z^{\dagger} (z)\Sigma_i^{-1} J_z(z) + (J_{\overline{z}}^{\dagger} (z) \Sigma_i^{-1} J_{\overline{z}}(z))^T  \succeq 0 ,
\end{align}
which gives the result, since therefore $\Psi_i / \rho^2 + \log a_i$ must be plurisubharmonic too. Here, we have decomposed $\Sigma^{-1} = \Gamma^{\dagger} \Gamma$, which is a square root decomposition and exists since $\Sigma^{-1}$ is positive semi-definite.

\noindent $ \square $

\vspace{2mm}

\noindent \textit{Remark} (non-rigorous, no proof provided). It can be noted that the potentials $K_1 = \rho^2 \log ( \sum_i \exp \{ - \Psi_i / \rho^2 \} )$, $K_2 = - \rho^2 \log ( \sum_i \exp \{ - \Psi_i / \rho^2 \} ) $ are not (necessarily) plurisubharmonic.

\vspace{2mm}

\noindent \textbf{Theorem 3.} Let $\tilde{h}$ be the Fisher information metric as in Theorem 1, and let $h$ be the Kähler-potential induced mertric as of equation \ref{eqn:Kähler_metric}. Moreover, suppose
\begin{align}
\partial v \approx M \Sigma^{-1} v ,
\end{align}
for some matrix $M$, where $v_i : = x(z) - \mu_i$. In particular, we have checked this empirically, and this is reasonable. Then if $x(z) \approx \mu_i$ for some $i$, there exists a constant $\rho$
\begin{align}
\tilde{h} \propto h .
\end{align}

\vspace{2mm}

\noindent \textit{Proof.} By Lemma 1, it follows for some $j$
\begin{align}
\EX_{w} [ \partial_{\alpha} \partial_{\overline{\beta}} \Psi] \approx (\partial_{\alpha}\mu_j)^{\dagger} \Sigma_j^{-1} (\partial_{\overline{\beta}} \mu_j) + (\partial_{\overline{\beta}}\mu_j)^{\dagger} \Sigma_j^{-1} (\partial_{\alpha} \mu_j) . 
\end{align}
Observe the other terms scale quadratically in $\Sigma^{-1}$, i.e.
\begin{align}
\text{Tr}\Bigg( \Sigma^{-1} (\partial_{\alpha} \Sigma) \Sigma^{-1} ( \partial_{\overline{\beta}}  \Sigma )\Bigg) \sim \mathcal{O}\Big(  ||\Sigma^{-1}||^2  \Big), \ \ \ \ \ 
\text{Cov}(\partial_{\alpha} \Psi, \partial_{\overline{\beta}} \Psi) \sim \mathcal{O}\Big( ||\Sigma^{-1}||^2 \Big ).
\end{align}
In particular, the second equation follows since
\begin{align}
\partial_{\alpha} \Psi_i & = (\partial_{\alpha} v)^{\dagger} \Sigma_j^{-1} v + v^{\dagger} \Sigma_j^{-1} (\partial_{\alpha} v)
\\
& \approx (M \Sigma_j^{-1} v)^{\dagger} \Sigma_j^{-1} v + v^{\dagger} \Sigma_j^{-1} (M \Sigma_j^{-1} v)
\\
& = v^{\dagger} ( \Sigma_j^{-1} M^{\dagger} \Sigma_j^{-1} +\Sigma_j^{-1} M \Sigma_j^{-1} ) v  =  v^{\dagger} P v . 
\end{align}
Again, there exists a weight, or weights, sufficiently large so that one, or multiple, $j \in I$ dominate. It is the case that $\Sigma_{j \in I}^{-1} \approx \Sigma(z)$. Thus,
\begin{align}
\text{Cov}(\partial_{\alpha} \Psi, \partial_{\overline{\beta}} \Psi) \approx \text{Cov}( v^{\dagger} P v, v^{\dagger} P v )  \sim \text{Tr}(P \Sigma P \Sigma)  \propto \mathcal{O}(||\Sigma^{-1}||^2) .
\end{align}
This is because $P \sim (\Sigma^{-1})^2$. The real version of covariance formula above is found in \cite{petersen2012matrix} on page 43. Here, we have noted $v \sim \mathcal{N}(0,\Sigma)$.

\noindent $ \square $

\vspace{2mm}

\noindent There is a nontrivial dilemma that we require $P$ to scale in some regard to $\partial_{\alpha} \Sigma, \partial_{\overline{\beta}} \Sigma$. Instead of presenting empirical evidence of this, we refer to our derivation in \ref{app:appendix_a}, which has theoretical support.

\vspace{10mm}

\vspace{-10mm}
\begin{figure}
  \vspace{0mm}
  \centering
  \includegraphics[scale=0.55]{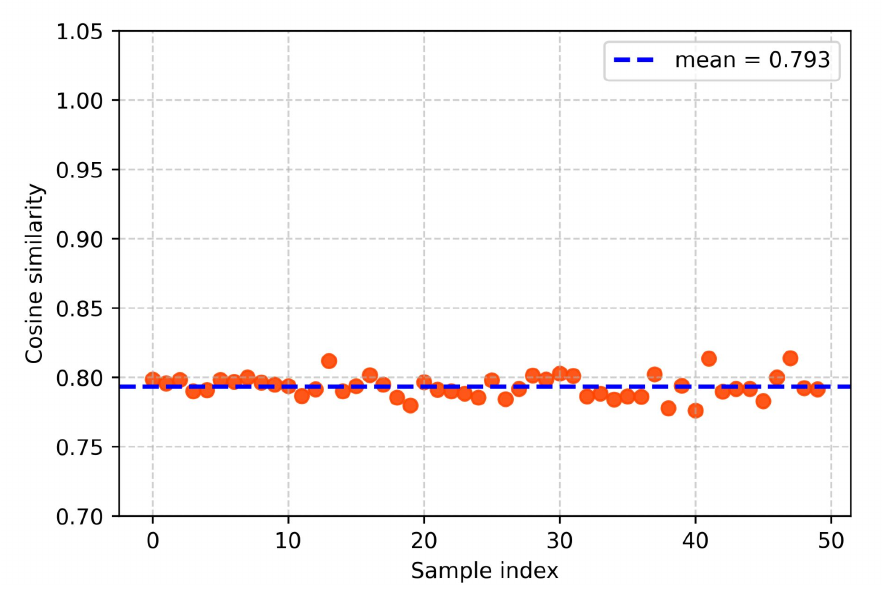}
  \caption{We illustrate the condition as in Theorem 3, $\partial v \approx M \Sigma^{-1} v$, is reasonable through a cosine similarity. In particular, a similarity of $+1\in [ -1, +1]$ is most desirable. Ours is 0.793.}
  \label{fig:method}
\end{figure}

\subsection{Inference}
\label{sec:sampling}

It is a known fact that there is a volume element equivalence
\begin{align}
dV_g = \sqrt{\text{det}(g)} \lambda_{2d} = |\text{det}(h)| \lambda_{2d} .
\end{align}
In particular, using the (1,1)-form, it is known \cite{stackexchange:4428928} (see \cite{Zheng_2017}, \cite{mandolesi2025hermitiangeometrycomplexmultivectors} for other relevant literature)
\begin{align}
\omega_p^d = d! 
\frac{i^d}{2^d} \zeta^1 \wedge \overline{\zeta}^1 \wedge \hdots \wedge \zeta^d \wedge \overline{\zeta}^d, \ \ \ \ \ \zeta^i \wedge \overline{\zeta}^i = -2 i \chi^i \wedge v^i ,
\end{align}
and we get $\omega_p^d = d! \text{Vol}_g$, but since $\omega^d = d! \text{det}(h)(i/2)^d dz^1 \wedge d\overline{z}^1 \wedge \hdots$, we have the result.

\vspace{2mm}

\noindent It is important to distinguish that sampling according the intrinsic \textit{decoder} geometry is a less meaningful task than sampling according to the \textit{encoder} geometry as in \cite{chadebec2022geometricperspectivevariationalautoencoders}. The decoder metric does not characterize the probability distributions of the model with respect to generative inference. The decoder geometry measures latent perturbations into ambient pixel-space changes. It captures geometry of representations, not geometry of inference.

\vspace{2mm}

\noindent When working with our metrics, we will generally consider admissible metrics closely related to \ref{eqn:fisher_metric} and our metric sufficiently close (up to constants)
\begin{align}
h(z) \in \Bigg\{ h \ \ \Bigg| \  \  1 \approx \frac{  \text{det}(h) }{ \text{det}  (  \partial \overline{\partial} \  \text{KL}( p(x|z) \parallel p(x|z') )  \Big|_{z'=  z}  )        } \sim    \text{det} (   \EX [ \EX [ \partial \overline{\partial} \Psi] + \frac{1}{\rho^2} \sum \text{Cov}(\partial \Psi, \overline{\partial} \Psi)    ] )      \Bigg\}  .
\end{align}
We have used $\sim$ to denote dependence, not necessarily proportionality. The determinant is real since $h$ is a valid Hermitian metric, and so
\begin{align}
h \in \Gamma(\text{Herm}^+(T^{1,0} M )) \cap \mathcal{A}_{\delta}, \ \ \ \mathcal{A}_{\delta} = \{ \ h \ | \ \text{det} \ H_{\Psi}(z)  > \EX_{z \sim \text{uniform}}[h(z)], \EX_{z \sim \text{uniform}} [H_{\text{KL}}(z)] \ \}  ,
\end{align}
where the inequality holds with high probability. More specifically, we can facilitate the metric computation via faster approximations. For example, oftentimes, we can sample the diagonal only for computational benefit, and so
\begin{align}
\text{det} (    \EX_{w} [ \partial \overline{\partial} \Psi] + \frac{1}{\rho^2} \text{Cov}(\partial \Psi, \overline{\partial} \Psi) )
& \propto \prod_{\alpha=1}^d \Big[ \EX_{w} [ \partial_{\alpha} \partial_{\overline{\alpha}} \Psi] + \frac{1}{\rho^2} \text{Cov}(\partial_{\alpha} \Psi, \partial_{\overline{\alpha}} \Psi) \Big]
\\
& = \prod_{\alpha=1}^d  \overline{ [\EX_{w} [ \partial_{\alpha} \partial_{\overline{\alpha}} \Psi  ] + \frac{1}{\rho^2} \text{Cov}(  \partial_{\alpha} \Psi ,  \partial_{\overline{\alpha}} \Psi ) ]} .
\end{align}
Our primary experiments will be done twofold. We will add a regularization / penalty loss, and so our loss function will take the form with sampling
\begin{align}
\label{eqn:loss_metric}
& L = L_{\text{reconst}} + \EX \Big[ \beta \text{KL} ( q(z | x) \parallel \mathcal{CN}(0,I) ) + \gamma \EX_q \Big[ \frac{1}{2} z^{\dagger} z - \frac{1}{2} \log \text{det} ( h(z) ) \Big]  \Big] .
\end{align}
The sampling is done at inference, so our metric surrogate is used in both training and inference. We make a few remarks. \noindent This loss encourages nondegeneracy ($h \approx 0$ will blow up), and extreme curvature ($h$ large will blow up negatively, but $z^{\dagger} z$ counteracts this). This loss helps mitigate ill-conditioning with the decoder metric. Moreover, we previously discussed advantages of this loss in section \ref{sec:proposed_metric}. In particular, our proposed softmax-type expectation rewards distance, and so does this loss. In particular, this loss function especially supports our methods.

\vspace{2mm}

\noindent \cite{nazari2023geometricautoencodersdecode} proposes penalizing the induced metric determinant, but rather its variance in a non-variational setting, and so no sampling is done in accordance with the penalty. \cite{lee2022regularized} presents isometric representation learning with a regularization penalty. It is nontrivial to compute the metric $h$ as part of the training procedure. Our solution to this is compute it every $K$ iterations and to attach the fixed, stored gradient to the gradient descent over iteration.

\begin{algorithm}
\caption{Metric sampling}\label{alg:diverse_posterior}
Take $(\mu,\sigma) = \text{encoder}(x)$, 
collect $\mu$, $\log\sigma^2$, set $\sigma = \exp \{ \frac{1}{2} \text{Re}( \log \sigma^2) \}$;
\\
\While{\text{Overdrawing samples}}{
  Sample batch, draw $\varepsilon \sim \frac{1}{\sqrt{2}} \mathcal{N}(0,I) + \frac{i}{\sqrt{2}} \mathcal{N}(0,I)$ \;
  Form generative candidates $z = \mu_i + \text{jitter} \cdot \sigma_i \odot \varepsilon$\;
  \ForEach{$z_j$ in $\{z_i\}_i$}{
    $\ell_j \leftarrow \log\det(h(z_j))$ \ \text{(clamp if necessary; do not clamp so that it is constant; normalize w/ \textit{accumulated} mean)}\;
    $w_j \leftarrow -\alpha\,\ell_j^2 - \lambda||z_j||^2$\;
  }
}

$\text{logits} \leftarrow (w - \overline{w}) / \text{temperature}$\;
$p \leftarrow \mathrm{softmax}(\text{logits})$\;
Sample indices from multinomial $p$ without replacement, keep $z_{\text{generative}}$

\end{algorithm}

\vspace{2mm}

\noindent Note that this procedure has connections to Ricci curvature, since, as a form and tensor respectively \cite{Viaclovsky2018Japan},
\begin{align}
& \text{Ric} = -i \partial \overline{\partial} \log \text{det} (h) = - i \sum_{ij} \frac{\partial^2 \log \text{det}(h)}{\partial z^i \partial \overline{z}^j} dz^i \wedge d\overline{z}^j, \text{Ric}_{ij} = - \partial_i \overline{\partial}_j \log \text{det} (h) = - \frac{\partial^2 \log \text{det}(h)}{\partial z^i \partial \overline{z}^j} ,
\end{align}
thus our regularization has an enhanced reformulation with notation $A=i A_{i \overline{j}} dz^i \wedge d\overline{z}^j$
\begin{align}
\text{L}_{\text{regularization enhanced}} & = \gamma \EX_{x \sim p_{\text{training}}} \EX_q \Big[ \ \Big| \Big| \frac{i}{2} \partial \overline{\partial} z^{\dagger} z + \frac{1}{2}  \text{Ric} \Big| \Big|_F \ \Big]
\\
& = \gamma \EX_{x \sim p_{\text{training}}} \EX_q \Big[ \sqrt{ \text{Tr} \Big( ( \frac{i}{2} I + \frac{1}{2}  \text{Ric} ) ( \frac{i}{2} I + \frac{1}{2}  \text{Ric} )^{\dagger} \Big) } \Big] ,
\end{align}
(note the cyclic property of the trace) using a gradient-enhanced approach as of \cite{Yu_2022}, which is a classical technique in physics-informed neural networks (PINNs).

\begin{table}[!htbp]
\caption{We illustrate outlier nearest neighbor distance (in the feature space) scores using a baseline complex VAE. on our geometric Kähler training and sampling. All sampling is done with the aggregate posterior. Hyperparameter $\beta$ to scale the KL divergence is held constant here among ours and the baseline. Lower is better. 9xth denotes percentile, and $\%$ refers to what percent of data is outliers. The MNIST experiment is without metric normalization, and the CIFAR experiment is with normalization (subtract accumulated mean).}
\label{tab:method_results_table}
\centering
\renewcommand{\arraystretch}{1.2}%
\footnotesize
\begin{tabular}{
  >{\raggedright\arraybackslash}p{5.5cm}
  >{\centering\arraybackslash}p{2.5cm}
  >{\centering\arraybackslash}p{2.5cm}
}
\toprule
\multicolumn{1}{c}{Feature nearest neighbor distance[$\downarrow$]} & \multicolumn{1}{c}{MNIST} & \multicolumn{1}{c}{CIFAR-10} \\
\midrule
\text{Among training data, 95th outlier distance} & {$0.408$} & {$1.277$}  \\
\text{Among training data, 99th  outlier distance} & {$0.467$} & {$1.531$} 
\\
\text{Complex VAE, mean dist.} & {$0.287$} & {$1.004$}  \\
\text{Complex VAE, 95th  outlier distance} & {$0.476$} & {$1.478$}  \\
\text{Complex VAE, 99th  outlier distance} & {$0.562$} & {$1.689$}  \\
\text{Complex VAE, 95th \% of outliers} & {$0.120$} & {$0.170$}  \\
\text{Complex VAE, 99th \% of outliers} & {$0.055$} & {$0.045$}  \\
\rowcolor{Orange!15} \text{Kähler sampling, mean dist.} & {$0.273$} & {$0.865$}  \\
\rowcolor{Orange!15} \text{Kähler sampling, 95th  outlier distance} & {$0.406$} & {$1.248$}  \\
\rowcolor{Orange!15} \text{Kähler sampling, 99th  outlier distance} & {$0.434$} & {$1.325$}  \\
\rowcolor{Orange!15} \text{Kähler sampling, 95th \%  of outliers} & {$0.050$} & {$0.045$}  \\
\rowcolor{Orange!15} \text{Kähler sampling, 99th \% of outliers} & {$0.005$} & {$0.005$}  \\
\addlinespace
\bottomrule
\end{tabular}
\end{table}

\section{Conclusions}

We have developed two Kähler potentials whose complex Hessian $\partial \overline{\partial}$ depends on the decoded Fisher information metric, one exact and one a proxy. Our approximate metric remains a valid Kähler potential $K$, and it is efficient because the computational burden is displaced from automatic differentiation to an expectation. This expectation is also more efficient than using a nearest neighbor algorithm. The efficiency matters because constructing a metric of the latent space and sampling from this metric is nontrivial. Our work is primarily interesting because the intersection of complex geometry and ML is a largely unexplored area.

\section{Acknowledgments}

I gratefully acknowledge financial support from Purdue University Department of Mathematics.

\bibliographystyle{plainnat}
\bibliography{bibliography}

\appendix

\newpage

\section{Connections between our Kähler potential and the Fisher metric with analytic equivalence}
\label{app:appendix_a}

In the primary section of our work earlier, we took
\begin{align}
x \sim \mathcal{CN}(x;\mu_i, \Sigma_i) .
\end{align}
In particular, our decoded output is closed to a true output $\mu_i$ that is previously sampled. Also, in that section, the weights our organized so that a particular, or multiple, $i$ dominate, thus the log of the sum of the exponentials is heavily skewed in favor of certain $i$. In practice, the weights with the softmax mean that this work is analogous to a single exponential in the sum. 

\vspace{2mm}

\noindent In this section, we will treat $\mu(z)$ as the sampled image, thus we will treat $\mu(z) \approx x$. Under sufficient closeness, it is true still true that $\mu(z) \approx \mu_i, \Sigma(z) \approx \Sigma_i$. With this framework, we can reconcile the differences between these two settings, one discrete and one continuous.

\vspace{2mm}

\noindent Let us take
\begin{align}
p(x |z) &= \mathcal{CN}(\mu(z), \Sigma(z))  ,
\end{align}
which is consistent with \cite{kingma2022autoencodingvariationalbayes} \cite{dorta2018trainingvaesstructuredresiduals}, and let us define
\begin{align}
\Psi(z; x) &= -\rho^2 \log p(x | z)
= \rho^2 \big( \log \det \Sigma + (x - \mu)^\dagger \Sigma^{-1} (x - \mu) \big) + \text{constant} .
\end{align}
Let $v := x - \mu(z)$. Since we will always take $\partial_{\alpha} \partial_{\overline{\beta}} \Gamma= 0$, so
\begin{align}
\partial_{\alpha} \Psi
& = \rho^2 \Big(
\mathrm{Tr}(\Sigma^{-1} \partial_{\alpha} \Sigma) - v^\dagger \Sigma^{-1} \partial_{\alpha} \mu - (\partial_{\alpha} \mu)^\dagger \Sigma^{-1} v
- v^\dagger \Sigma^{-1} (\partial_{\alpha} \Sigma) \Sigma^{-1} v \Big),
\end{align}
and similarly for $ \partial_{\overline{\beta}} \Psi$.

\vspace{2mm}

\noindent For the complex Gaussian, we have
\begin{align}
\mathbb{E}[v] = 0, \ \ \ \ \ \mathbb{E}[v v^\dagger] = \Sigma,  \ \ \ \ \ 
\mathrm{Cov}(v^\dagger A v, \, v^\dagger B v) = \mathrm{Tr}(A \Sigma B \Sigma),
\end{align}
for any fixed matrices $A,B$. Thus, 
\begin{align}
\partial_{\overline{\beta}} \partial_{\alpha} \Psi
&= \rho^2 \Big[
\partial_{\overline{\beta}} \Big( \mathrm{Tr}(\Sigma^{-1}\partial_\alpha\Sigma) \Big) -\partial_{\overline{\beta}} \Big( v^{\dagger} \Sigma^{-1} \partial_{\alpha} \mu \Big) - \partial_{\overline{\beta}} \Big( ( \partial_{\alpha} \mu)^{\dagger} \Sigma^{-1} v \Big)
- \partial_{\overline{\beta}} \Big( v^\dagger\Sigma^{-1}(\partial_\alpha\Sigma)\Sigma^{-1}v \Big)
\Big].
\end{align}
and notice
\begin{gather}
\partial_{\overline{\beta}} \Big( \mathrm{Tr}(\Sigma^{-1}\partial_\alpha\Sigma) \Big)
= \mathrm{Tr} \Big(
(\partial_{\overline{\beta}}\Sigma^{-1})\partial_\alpha\Sigma
+ \Sigma^{-1}\partial_{\overline{\beta}}\partial_{\alpha}  \Sigma
\Big),
\\
\partial_{\overline{\beta}} \Big( v^{\dagger} \Sigma^{-1} \partial_{\alpha} \mu \Big) = -(\partial_{\overline{\beta}} \mu)^{\dagger} \Sigma^{-1} (\partial_{\alpha} \mu ) + v^{\dagger}\partial_{\overline{\beta}} \Sigma^{-1} \partial_{\alpha } \mu 
\\
\partial_{\overline{\beta}} \Big( ( \partial_{\alpha} \mu)^{\dagger} \Sigma^{-1} v \Big) = -(\partial_{\alpha} \mu)^{\dagger} \Sigma^{-1} \partial_{\overline{\beta}} \mu + (\partial_{\alpha} \mu)^{\dagger} \partial_{\overline{\beta}} \Sigma^{-1}  v
\\
\partial_{\overline{\beta}} \Big( v^\dagger\Sigma^{-1}(\partial_\alpha\Sigma)\Sigma^{-1}v \Big) 
= v^\dagger(\partial_{\overline{\beta}}\Sigma^{-1})(\partial_\alpha\Sigma)\Sigma^{-1}v
+ v^\dagger\Sigma^{-1}(\partial_\alpha\Sigma)(\partial_{\overline{\beta}}\Sigma^{-1})v
+ v^\dagger\Sigma^{-1}(\partial_{\overline{\beta}}\partial_\alpha\Sigma)\Sigma^{-1}v.
\end{gather}
The last equation is missing terms, which go to zero after expectation. We will use the identities
\begin{align}
\partial_{\gamma}\Sigma^{-1} = -\Sigma^{-1}(\partial_{\gamma}\Sigma)\Sigma^{-1}, \ \ \ 
\mathbb{E}[\,v^\dagger A v\,] = \mathrm{Tr}(A\Sigma), \ \ \ \EX[ v^{\dagger} \Sigma^{-1} \partial_{\gamma} \mu ] = 0.
\end{align}
thus, after annihilating terms,
\begin{align}
\mathbb{E}[\partial_{\alpha} \partial_{\overline{\beta}} \Psi] &  =  \rho^2 \EX \Bigg[ \text{Tr}(- \Sigma^{-1} \partial_{\overline{\beta}} \Sigma \Sigma^{-1} \partial_{\alpha} \Sigma ) + (\partial_{\overline{\beta}} \mu)^{\dagger} \Sigma^{-1} (\partial_{\alpha} \mu)  +  (\partial_{\alpha} \mu)^{\dagger} \Sigma^{-1} (\partial_{\overline{\beta}} \mu)
\\
& \ \ \ \ \ \ \ \ \ \ \ \ \ \ \ \ + v^{\dagger} \Sigma^{-1} \partial_{\overline{\beta}} \Sigma \Sigma^{-1} \partial_{\alpha} \Sigma \Sigma^{-1} v + v^{\dagger} \Sigma^{-1} \partial_{\alpha} \Sigma \Sigma^{-1} \partial_{\overline{\beta}} \Sigma \Sigma^{-1} v      \Bigg]
\\
& =  \rho^2 \Bigg( 2 \text{Re} (\partial_{\alpha} \mu)^{\dagger} \Sigma^{-1} (\partial_{\overline{\beta}} \mu ) + \mathrm{Tr} \big( \Sigma^{-1} (\partial_{\alpha} \Sigma) \Sigma^{-1} (\partial_{\overline{\beta}} \Sigma) \big) \Bigg).
\end{align}
We have used invariance of the trace under circular shifts. Also,
\begin{align}
& \mathrm{Cov}(\partial_{\alpha} \Psi, \partial_{\overline{\beta}} \Psi)
\\
& = \EX [ \partial_{\alpha} \Psi \partial_{\overline{\beta}} \Psi ] - \underbrace{ \EX [  \partial_{\alpha} \Psi  ] \EX [ \partial_{\overline{\beta}} \Psi  ] }_{=0} = \EX [ \partial_{\alpha} \Psi \partial_{\overline{\beta}} \Psi ]
\\
& = \EX \Bigg[ \rho^2 \Big(
\mathrm{Tr}(\Sigma^{-1} \partial_{\alpha} \Sigma) - v^\dagger \Sigma^{-1} \partial_{\alpha} \mu - (\partial_{\alpha} \mu)^\dagger \Sigma^{-1} v
- v^\dagger \Sigma^{-1} (\partial_{\alpha} \Sigma) \Sigma^{-1} v \Big) 
\\
& \ \ \ \ \ \ \ \ \ \ \times \ \ \ \rho^2 \Big(
\mathrm{Tr}(\Sigma^{-1} \partial_{\overline{\beta}} \Sigma) - v^\dagger \Sigma^{-1} \partial_{\overline{\beta}} \mu - (\partial_{\overline{\beta}} \mu)^\dagger \Sigma^{-1} v
- v^\dagger \Sigma^{-1} (\partial_{\overline{\beta}} \Sigma) \Sigma^{-1} v \Big) \Bigg]
\\
& = \rho^4 \Bigg( 2 \EX \Big[ (v^\dagger \Sigma^{-1} \partial_{\alpha} \mu) \
\big((\partial_{\overline{\beta}} \mu)^\dagger \Sigma^{-1} v\big) \Big] 
\\
& \ \ \ \ \ \ \ \ \ \ \ \ \ + \EX \Bigg[ \Big(
\mathrm{Tr}(\Sigma^{-1} \partial_{\alpha} \Sigma)  
- v^\dagger \Sigma^{-1} (\partial_{\alpha} \Sigma) \Sigma^{-1} v \Big)  \Big(
\mathrm{Tr}(\Sigma^{-1} \partial_{\overline{\beta}} \Sigma) 
- v^\dagger \Sigma^{-1} (\partial_{\overline{\beta}} \Sigma) \Sigma^{-1} v \Big) \Bigg] \Bigg)
\\
& = \rho^4 \Bigg(  2 \text{Re} (\partial_{\alpha} \mu)^\dagger \Sigma^{-1} (\partial_{\overline{\beta}} \mu) + 
\mathrm{Tr}(\Sigma^{-1} \partial_{\alpha} \Sigma)   \mathrm{Tr}(\Sigma^{-1} \partial_{\overline{\beta}} \Sigma)  - \mathrm{Tr}(\Sigma^{-1} \partial_{\alpha} \Sigma) \EX \Big[ v^\dagger \Sigma^{-1} (\partial_{\overline{\beta}} \Sigma) \Sigma^{-1} v \Big] 
\\
& \ \ \ \ \ \ \ \ \ \ \ \ - \mathrm{Tr}(\Sigma^{-1} \partial_{\overline{\beta}} \Sigma) \EX \Big[ v^\dagger \Sigma^{-1} (\partial_{\alpha} \Sigma) \Sigma^{-1} v  \Big] + \EX \Big[ v^\dagger \Sigma^{-1} (\partial_{\alpha} \Sigma) \Sigma^{-1} v  v^\dagger \Sigma^{-1} (\partial_{\overline{\beta}} \Sigma) \Sigma^{-1} v  \Big] \Bigg)
\\
& = \rho^4 \Bigg( 2 \text{Re} (\partial_{\alpha} \mu)^\dagger \Sigma^{-1} (\partial_{\overline{\beta}} \mu)  -   
\mathrm{Tr}(\Sigma^{-1} \partial_{\alpha} \Sigma)   \mathrm{Tr}(\Sigma^{-1} \partial_{\overline{\beta}} \Sigma) + \EX \Big[ v^\dagger \Sigma^{-1} (\partial_{\alpha} \Sigma) \Sigma^{-1} v  v^\dagger \Sigma^{-1} (\partial_{\overline{\beta}} \Sigma) \Sigma^{-1} v  \Big] \Bigg)
\\
& = \rho^4 \Bigg( 2 \text{Re} (\partial_{\alpha} \mu)^\dagger \Sigma^{-1} (\partial_{\overline{\beta}} \mu) -   
\mathrm{Tr}(\Sigma^{-1} \partial_{\alpha} \Sigma)   \mathrm{Tr}(\Sigma^{-1} \partial_{\overline{\beta}} \Sigma) 
\\
& \ \ \ \ \ \ \ \ \ \ \ \ \ \ \ \ \ \ \ \ \ \ \ \ \ \ + \text{Tr} ( \Sigma^{-1} (\partial_{\alpha} \Sigma) ) \text{Tr} ( \Sigma^{-1} (\partial_{\overline{\beta}} \Sigma) ) + \text{Tr} \Big( \Sigma^{-1} \partial_{\alpha} \Sigma \Sigma^{-1} \partial_{\overline{\beta}} \Sigma \Big) \Bigg) 
\\
& = \rho^4 \Bigg(
 2 \text{Re} (\partial_{\alpha} \mu)^\dagger \Sigma^{-1} (\partial_{\overline{\beta}} \mu)
+ \mathrm{Tr} \Big( \Sigma^{-1} (\partial_{\alpha} \Sigma) \Sigma^{-1} (\partial_{\overline{\beta}} \Sigma) \Big)
\Bigg).
\end{align}
We have used $\EX[v] = 0$ and \cite{petersen2012matrix}
\begin{align}
\EX [ v^{\dagger} A v v^{\dagger} B v ] = \text{Tr}(A \Sigma) \text{Tr}(B \Sigma) + \text{Tr}(A \Sigma B \Sigma) .
\end{align}
Thus,
\begin{align}
\mathbb{E}[\partial_{\alpha} \partial_{\overline{\beta}} \Psi]
+ \frac{1}{\rho^2} \, \mathrm{Cov}(\partial_{\alpha} \Psi, \partial_{\overline{\beta}} \Psi)
&= \rho^2 \Big( 
4 \text{Re} (\partial_{\alpha} \mu)^\dagger \Sigma^{-1} (\partial_{\overline{\beta}} \mu)
+ 2\mathrm{Tr} \big( \Sigma^{-1} (\partial_{\alpha} \Sigma) \Sigma^{-1} (\partial_{\overline{\beta}} \Sigma) \big)
\Big)
\end{align}
exactly.

\vspace{2mm}

\noindent In practice, this is not the same object as derived in \ref{sec:proposed_metric} with negative values. For example, if we take what we have in the above, $\Psi = - \log p(x | z)$, we see
\begin{align}
\psi(z) & = - \rho^2 \log \int e^{- \Psi } d\mu  = - \rho^2 \log \int e^{\log p(x | z)} d\mu = - \rho^2 \int p(x | z) d\mu = -\rho^2 ,
\end{align}
thus
\begin{align}
\partial \overline{\partial} \psi(z) = 0 .
\end{align}

\noindent Let us examine the regime under the circumstance of sufficient closeness. Let us assume parameter closeness and $\gamma I \preceq \Sigma_i$. With a slight lack of rigor, consider the following
\begin{align}
& \text{KL} ( \mathcal{CN}(\mu(z), \Sigma(z)) \parallel \mathcal{CN}(\mu_i, \Sigma_i) ) 
\\
& \propto \text{Tr} ( \Sigma_i^{-1} \Sigma(z)) - \log \text{det} ( \Sigma_i^{-1} \Sigma(z) )  -d  + (\mu_i - \mu(z))^{\dagger} \Sigma_i^{-1} ( \mu_i - \mu(z) ) 
\\
& = \text{Tr} ( \Sigma_i^{-1} \Sigma(z) ) - \log \text{det} ( \Sigma_i^{-1} \Sigma(z) )  -d  + \Psi_i \Big|_{x = \mu(z)}
\\
& \leq |  \text{Tr} ( \Sigma_i^{-1} \Sigma(z) )  - d | + | \log \text{det} ( \Sigma_i^{-1} \Sigma(z) ) | + \frac{1}{\gamma} | | \mu_i - \mu(z) | |^2 
\\
& \leq \text{constant} \cdot \epsilon_{\Sigma}^2 + \text{constant} \cdot \epsilon_{\mu}^2
\end{align}
Under the Pisnker inequality, using the total variation norm (for signed measures, since densities are Radon-Nikodym derivatives of measures)
\begin{align}
\Bigg| \Bigg|  \mathcal{CN}(\mu(z), \Sigma(z)) - \mathcal{CN}(\mu_i, \Sigma_i) \Bigg| \Bigg|_{\text{TV}} & \leq \sqrt{ 2 \text{KL} (  \mathcal{CN}(\mu(z), \Sigma(z)) \parallel \mathcal{CN}(\mu_i, \Sigma_i)  ) }
\leq C \sqrt{\epsilon_{\Sigma}^2 + \epsilon_{\mu}^2 } .
\end{align}
We have defined the total variation norm as
\begin{align}
\sup_{\substack{
\phi \in C_c^1(\mathbb{R}^{2d}, \mathbb{R}^{2d}) \\
\|\phi\|_{L^\infty} \leq 1}}
\Bigg|
\int_{\mathbb{R}^{2d}}
\big( p_z(x) - p_i(x) \big) \
\operatorname{div}\phi(x) \ dx .
\Bigg|
\end{align}
In particular, this definition is meaningful due to an application of integration by parts and compact support, i.e. we implicitly measure the gradients on the densities. For example, by Green's first identity,
\begin{align}
\int_{K} ( \psi \nabla \cdot \Gamma + \Gamma \cdot \nabla \psi ) dV = \oint_{\partial K} \psi \Gamma \cdot n dS .
\end{align}
Therefore, where we have denoted $K$ the support,
\begin{align}
\int_{K}
\big( p_z(x) - p_i(x) \big)
\operatorname{div}\phi(x) dx
+ \int_K \nabla ( p_z(x) - p_i(x) ) \cdot \phi(x) dx  = 0 .
\end{align}

\section{Expectation and covariance relations}
\label{sec:exp_and_cov_relations}

In this section, we draw connections between the primary section of Section \ref{sec:proposed_metric} and Appendix \ref{app:appendix_a}.

\vspace{2mm}

\noindent We previously had
\begin{align}
& g_{\alpha \overline{\beta}} = \EX_{x | z} [ \partial_{\alpha} \partial_{\overline{\beta}} \Psi ]  + \frac{1}{\rho^2} \text{Cov}_{x | z}(\partial_{\alpha} \Psi, \partial_{\overline{\beta}} \Psi ) 
\\
& h_{\alpha \overline{\beta}} = \EX_{w(i)} [ \partial_{\alpha} \partial_{\overline{\beta}} \Psi_i ]  + \frac{1}{\rho^2} \text{Cov}_{w(i)}(\partial_{\alpha} \Psi_i, \partial_{\overline{\beta}} \Psi_i ) .
\end{align}
Let us take $i \sim w(i), x|z,i \sim p_i(\cdot | z)$. By total expectation,
\begin{align}
\EX_{i, x|z} [ \partial_{\alpha} \partial_{\overline{\beta}} \Psi(i,x;z) ] & = \EX_{i \sim w(i)} [ \EX_{x|z,i} [\partial_{\alpha} \partial_{\overline{\beta}} \Psi_i(x;z) ] ] 
\\
\text{Cov}_{i,x|z} ( \partial_{\alpha} \Psi(i,x;z), \partial_{\overline{\beta}} \Psi(i,x;z) ) & = \EX_{i \sim w(i)}  [ \text{Cov}_{x|z,i} ( \partial_{\alpha} \Psi_i(x;z), \partial_{\overline{\beta}} \Psi_i(x;z) ) ]
\\
& \ \ \ \ \ \ + \text{Cov}_{i \sim w(i)} ( \EX_{x|z,i} [ \partial_{\alpha} \Psi_i(x;z) ], \EX_{x|z,i} [ \partial_{\overline{\beta}} \Psi_i(x;z) ] ) .
\end{align}
Define the joint object
\begin{align}
G_{\alpha \overline{\beta}}(z) = \EX_{i, x|z} [ \partial_{\alpha} \partial_{\overline{\beta}} \Psi_i(x;z)] + \frac{1}{\rho^2} \text{Cov}_{i,x|z}(\partial_{\alpha} \Psi(i,x;z), \partial_{\overline{\beta}} \Psi(i,x;z) ) .
\end{align}
Thus,
\begin{align}
G_{\alpha \overline{\beta}}(z) & = \EX_{i \sim w(i) } \Bigg[ \EX_{x|z,i} [ \partial_{\alpha} \partial_{\overline{\beta}} \Psi_i(x;z) ] + \frac{1}{\rho^2} \text{Cov}_{x|z,i} (\partial_{\alpha} \Psi_i(x;z), \partial_{\overline{\beta}} \Psi_i(x;z)) \Bigg]
\\
& \ \ \ \ \ \ \ \ \ \ \ \ \ \ \ \ \ \ \ \ \ + \frac{1}{\rho^2} \text{Cov}_{i \sim w(i)} ( \EX_{x|z,i} [ \partial_{\alpha} \Psi_i(x;z) ], \EX_{x|z,i} [ \partial_{\overline{\beta}} \Psi_i(x;z) ] ) .
\end{align}
We conclude
\begin{align}
G_{\alpha \overline{\beta}}(z) & \propto \mathbb{E}_{w(i)} \Bigg[ \rho^2 \Big( 
    4 \text{Re} (\partial_{\alpha} \mu_i)^\dagger \Sigma_i^{-1} (\partial_{\overline{\beta}} \mu_i) 
    + 2\mathrm{Tr} \big( \Sigma_i^{-1} (\partial_{\alpha} \Sigma_i) \Sigma_i^{-1} (\partial_{\overline{\beta}} \Sigma_i) \big) 
\Big) \Bigg]  \\
& \ \ \ \ \ \ \ \ \ \ \ \ \ \ \ \ \ \ \ \ \ + \frac{1}{\rho^2} \mathrm{Cov}_{w(i)} \Big( \mathbb{E}_{x|z,i} [ \partial_{\alpha} \Psi_i ], \mathbb{E}_{x|z,i} [ \partial_{\overline{\beta}} \Psi_i ] \Big) .
\end{align}

\newpage

\section{Additional figures}
\label{app:additional_figures}

\vspace{20mm}

\begin{figure}[h!]
  \vspace{0mm}
  \centering
  \includegraphics[scale=0.7]{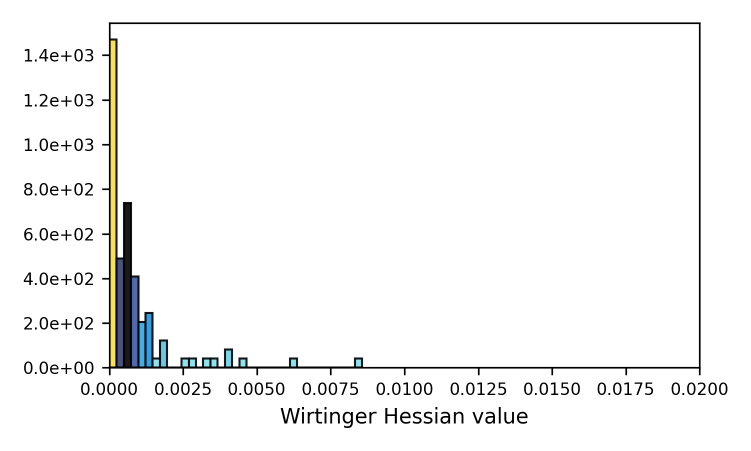}
  \caption{We plot 100 randomly sampled diagonal entries $\partial_{\alpha} \partial_{\overline{\alpha}}  (\text{decoder}(z).\texttt{mean}() )$ in a histogram. The mean is approximately $0.10$ on MNIST data, so the pixel data is nontrivial in value compares to its second order Wirtinger derivatives. The purpose of this plot is to convey the assumption $\partial_{\alpha} \partial_{\overline{\beta}} \mu(z) \approx 0$ is reasonable. We plot the absolute value of each component, and the max across two elements $\alpha$.}
  \label{fig:hii_hist}
\end{figure}

\begin{figure}[h!]
  \vspace{0mm}
  \centering
  \includegraphics[scale=0.7]{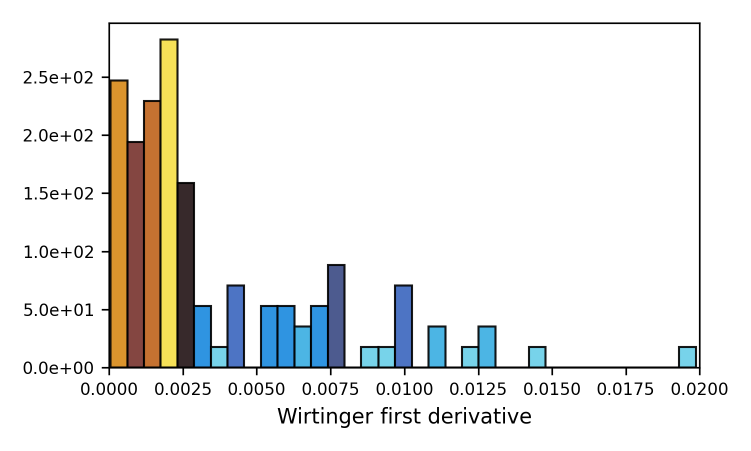}
  \caption{We plot 100 randomly sampled diagonal entries $\partial_{\alpha}   (\text{decoder}(z).\texttt{mean}() )$ in a histogram. The mean is approximately $0.10$ on MNIST data, so the pixel data is nontrivial in value compares to its second order Wirtinger derivatives. The purpose of this plot is to convey the first derivatives are much larger in magnitude on average with our architectures than the mixed second derivatives. Moreover, the plot with respect to $\overline{\alpha}$ is the same, thus the decoder is not holomorphic. We plot the absolute value of each component, and the max across two elements $\alpha$. The x-scale is consistent with Figure \ref{fig:hii_hist}. As a last remark, note that $\max |\partial_{\overline{z}}f - \overline{\partial_z f}|=0$, but $\max |\partial_z f - \partial_{\overline{z}} f| = 0.0281 $ on our real data.}
  \label{fig:first_deriv_hist}
\end{figure}

\begin{figure}[h!]
  \vspace{0mm}
  \centering
  \includegraphics[scale=0.45]{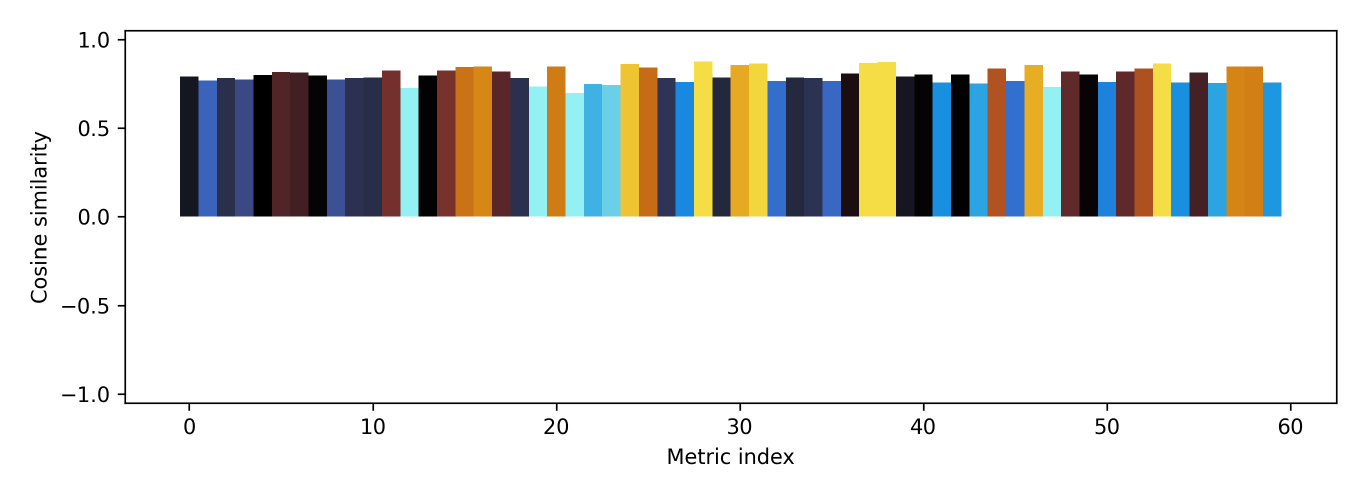}
  \caption{We present a cosine similarity between $\argmin_i \Psi_i$ and $\argmax_i \Psi_i$ on 50 posterior samples corresponding to  $2 \times \text{latent dim}$ indices (so one real, one imaginary) of our Kähler metric using pure weights of $w=1$ to replace the softmax function as of \ref{eqn:improved_weights}. This figure is created without normalization via subtracting the mean, thus they are more easily correlated. This figure was created using our MNIST experiment.}
  \label{fig:cos_similarity_psi}
\end{figure}

\begin{figure}[h!]
  \vspace{0mm}
  \centering
  \includegraphics[scale=0.45]{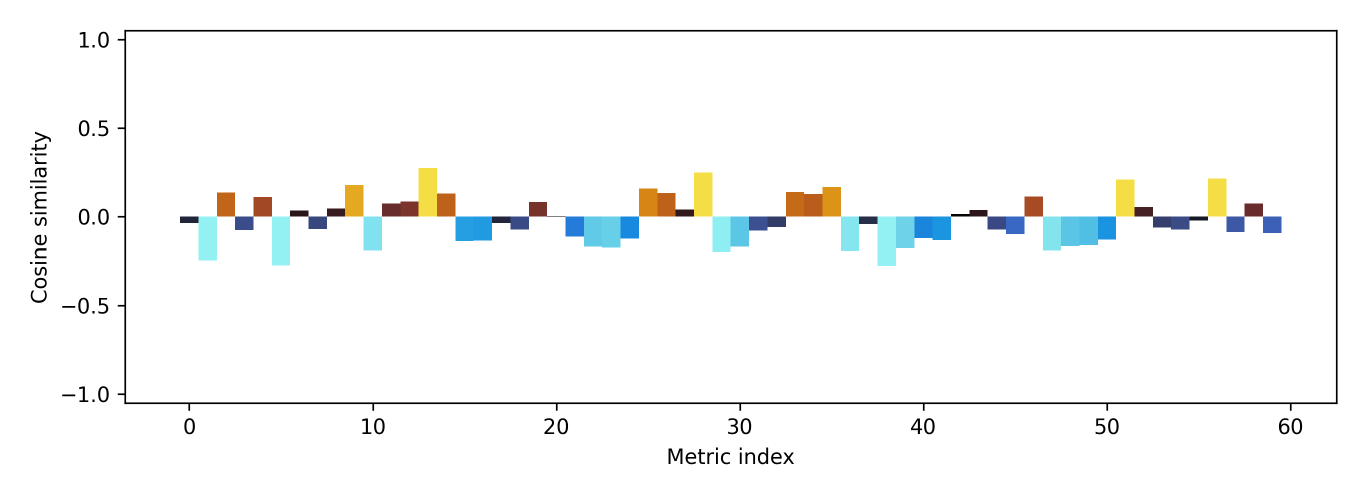}
  \caption{We present a cosine similarity between $\argmin_i \Psi_i$ and $\argmax_i \Psi_i$ on 50 posterior samples corresponding to  $2 \times \text{latent dim}$ indices (so one real, one imaginary) of our Kähler metric using pure weights of $w=1$ to replace the softmax function as of \ref{eqn:improved_weights}. This figure is created with normalization via subtracting the mean, thus we capture correlation due to metric differences. This figure does not imply sampling according to our metric is unreasonable due to a lack of similarity. Indeed, this figure demonstrates a lack of similarity in regard to proportional scaling (which is anticipated), but recall our loss function encourages a high metric along the distant regions.}
  \label{fig:cos_similarity_normalized}
\end{figure}

\begin{figure}
  \vspace{0mm}
  \centering
  \includegraphics[scale=0.25]{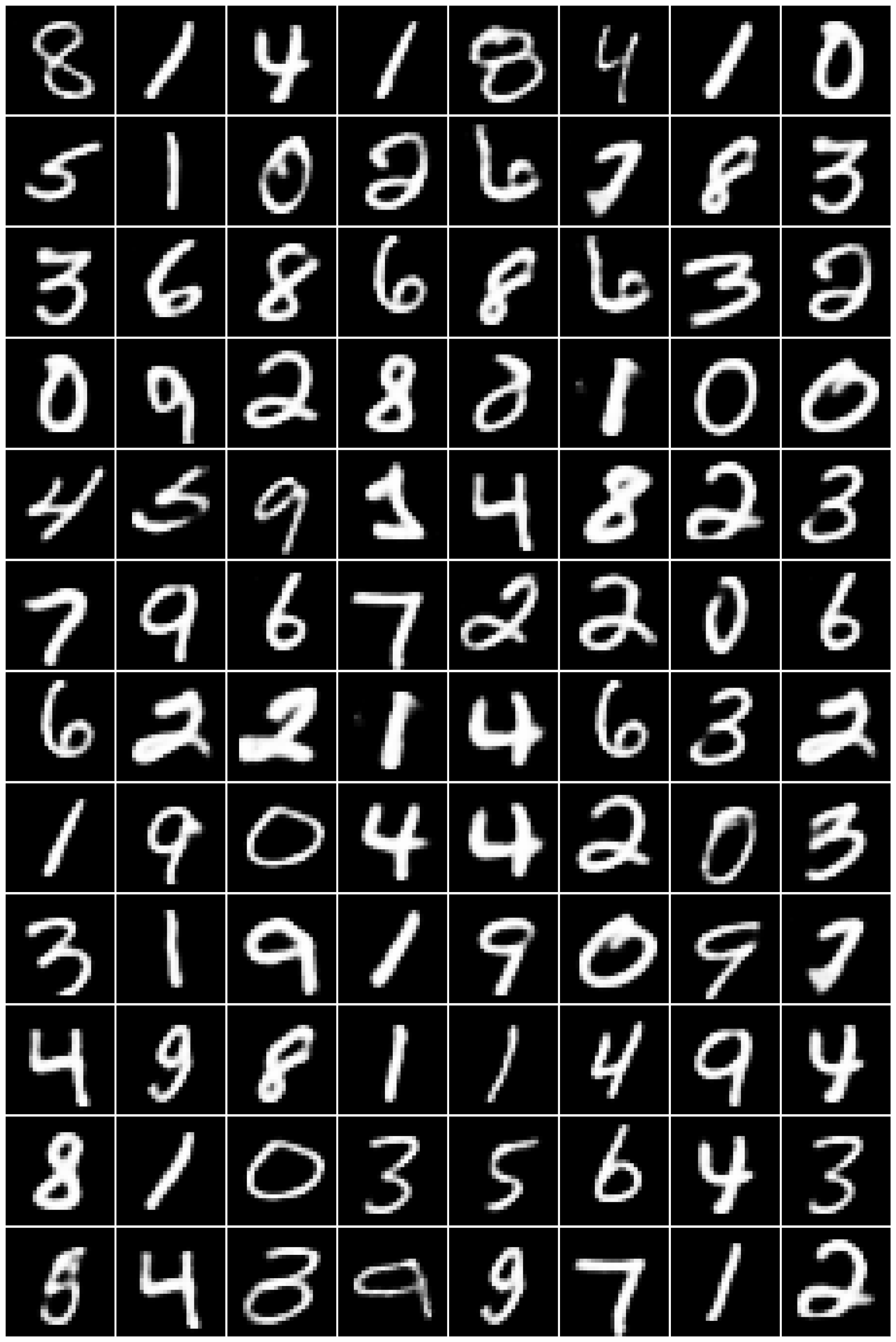}
  \caption{Uncurated generative samples from our Kähler-sampled VAE with posterior sampling with parameters corresponding to Table \ref{tab:method_results_table}. In particular, among this data, there are near zero outliers in the feature space, which is not true for a complex VAE with the same parameters (there are actually quite a few outliers for a complex VAE with the same setup; we refer to Table \ref{tab:method_results_table}). This comes at a cost of sample variation.}
  \label{fig:Kähler_mnist}
\end{figure}

\begin{figure}
  \vspace{0mm}
  \centering
  \includegraphics[scale=0.25]{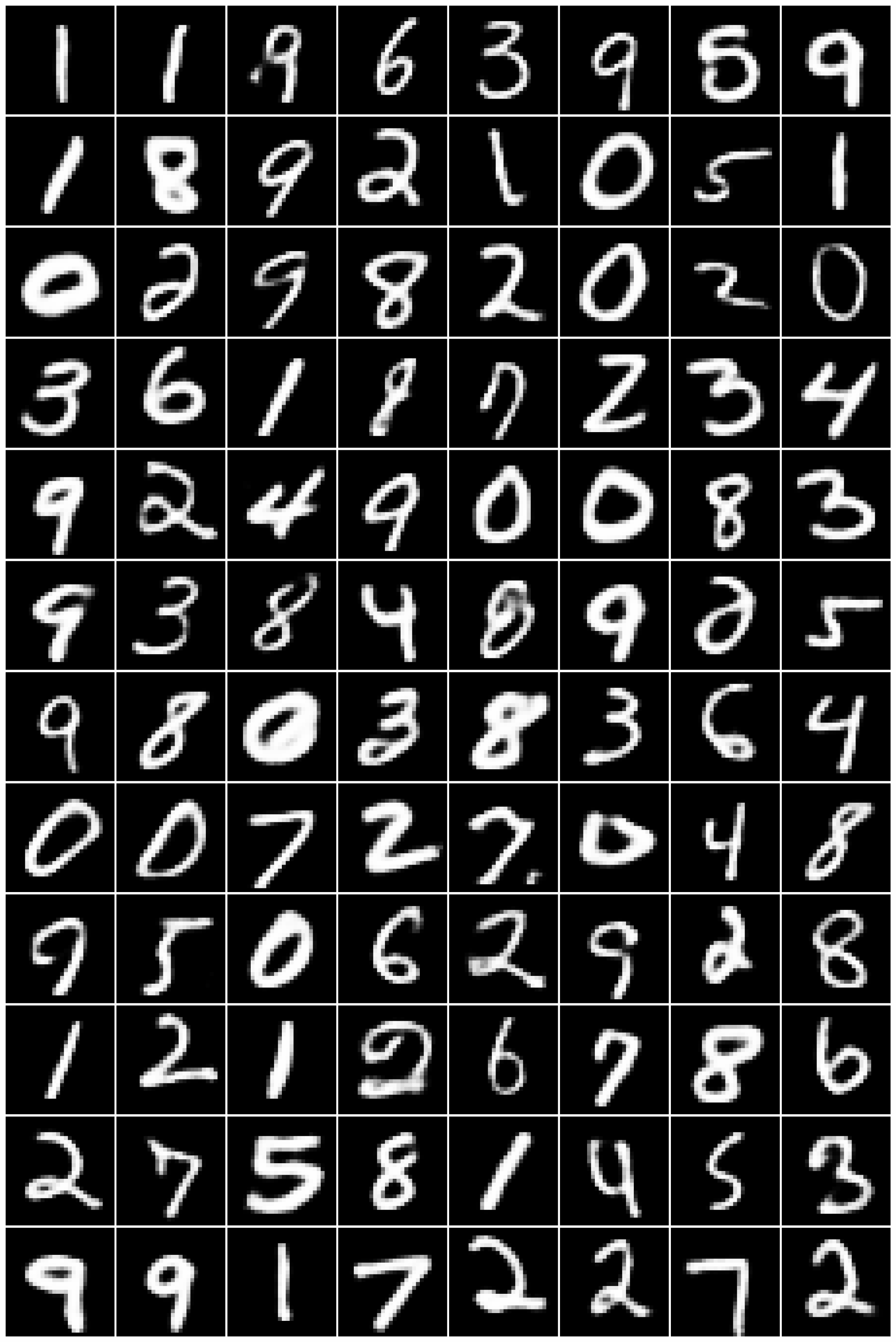}
  \caption{Uncurated generative samples from our baseline complex VAE with posterior sampling with parameters corresponding to Table \ref{tab:method_results_table}. Unlike our method, there do exist outliers in the feature space in this data.}
  \label{fig:baseline_mnist}
\end{figure}

\begin{figure}
  \vspace{0mm}
  \centering
  \includegraphics[scale=0.25]{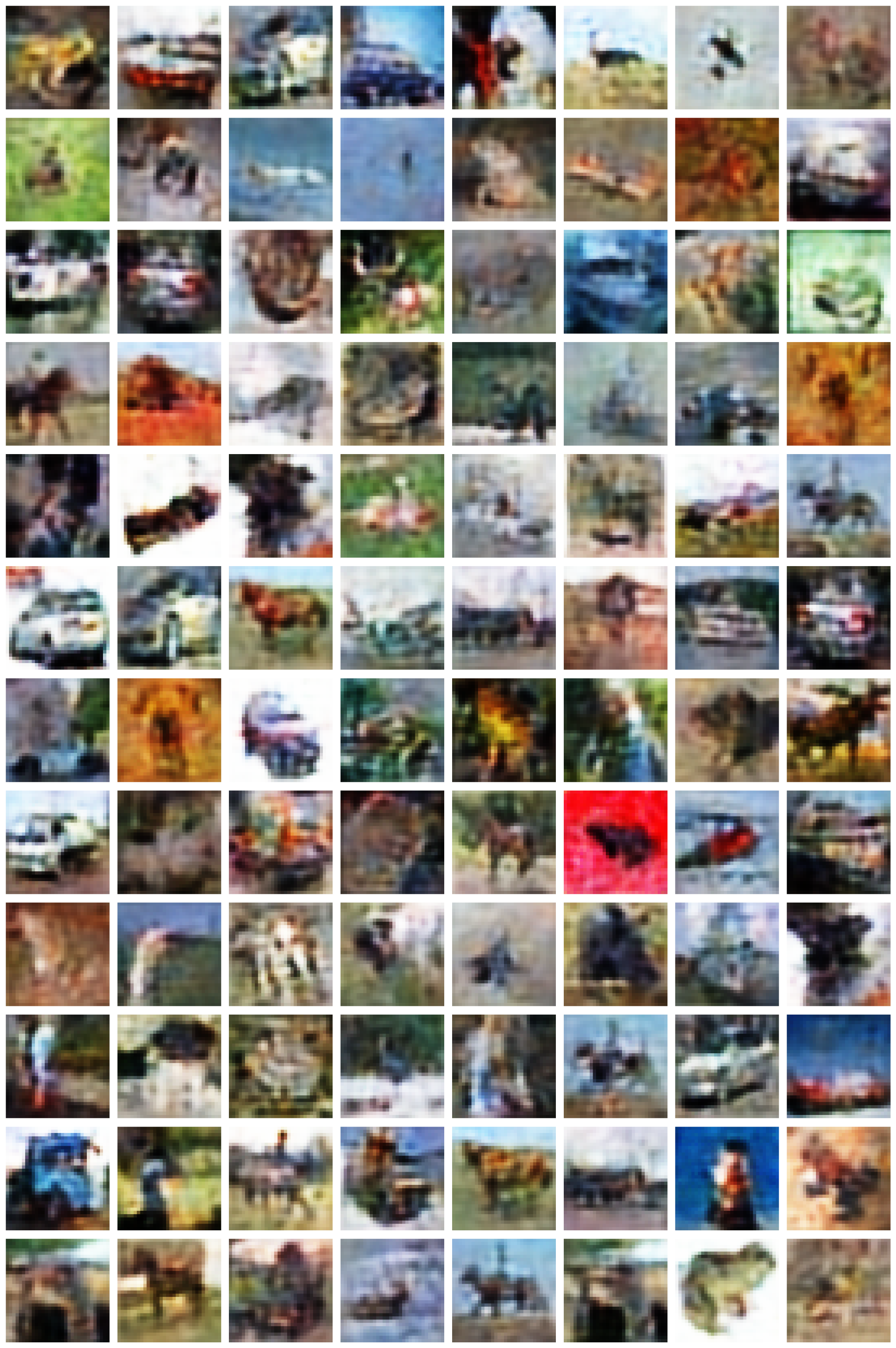}
  \caption{Uncurated generative samples from our Kähler-sampled VAE on our CIFAR-10 experiment with curvature-aware posterior sampling. We choose noise scaling parameter of $4$ here (meaning we take $\text{jitter} \cdot \epsilon \odot \sigma$, $\text{jitter}=4$ for our sampling), thus these results are consistent with the scenario of Table \ref{tab:method_results_table}.}
  \label{fig:cifar_sampling}
\end{figure}

\begin{figure}
  \vspace{0mm}
  \centering
  \includegraphics[scale=0.25]{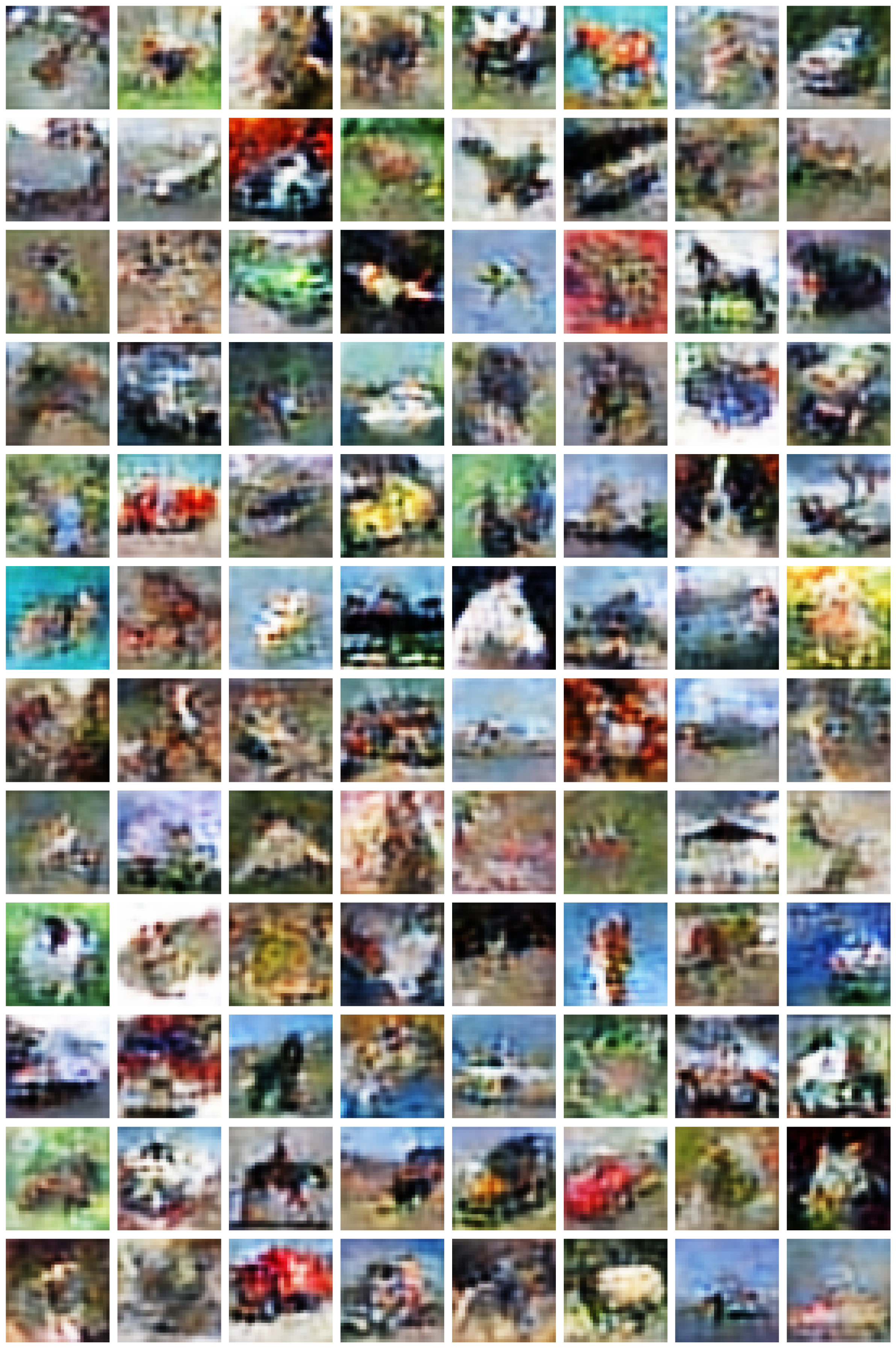}
  \caption{Uncurated generative samples from a complex VAE on our CIFAR-10 experiment. We choose noise scaling parameter of $4$ here (meaning we take $\text{jitter} \cdot \epsilon \odot \sigma$, $\text{jitter}=4$ for our sampling), thus these results are consistent with the scenario of Table \ref{tab:method_results_table}.}
  \label{fig:cifar_sampling}
\end{figure}

\begin{figure}
  \vspace{0mm}
  \centering
  \includegraphics[scale=0.25]{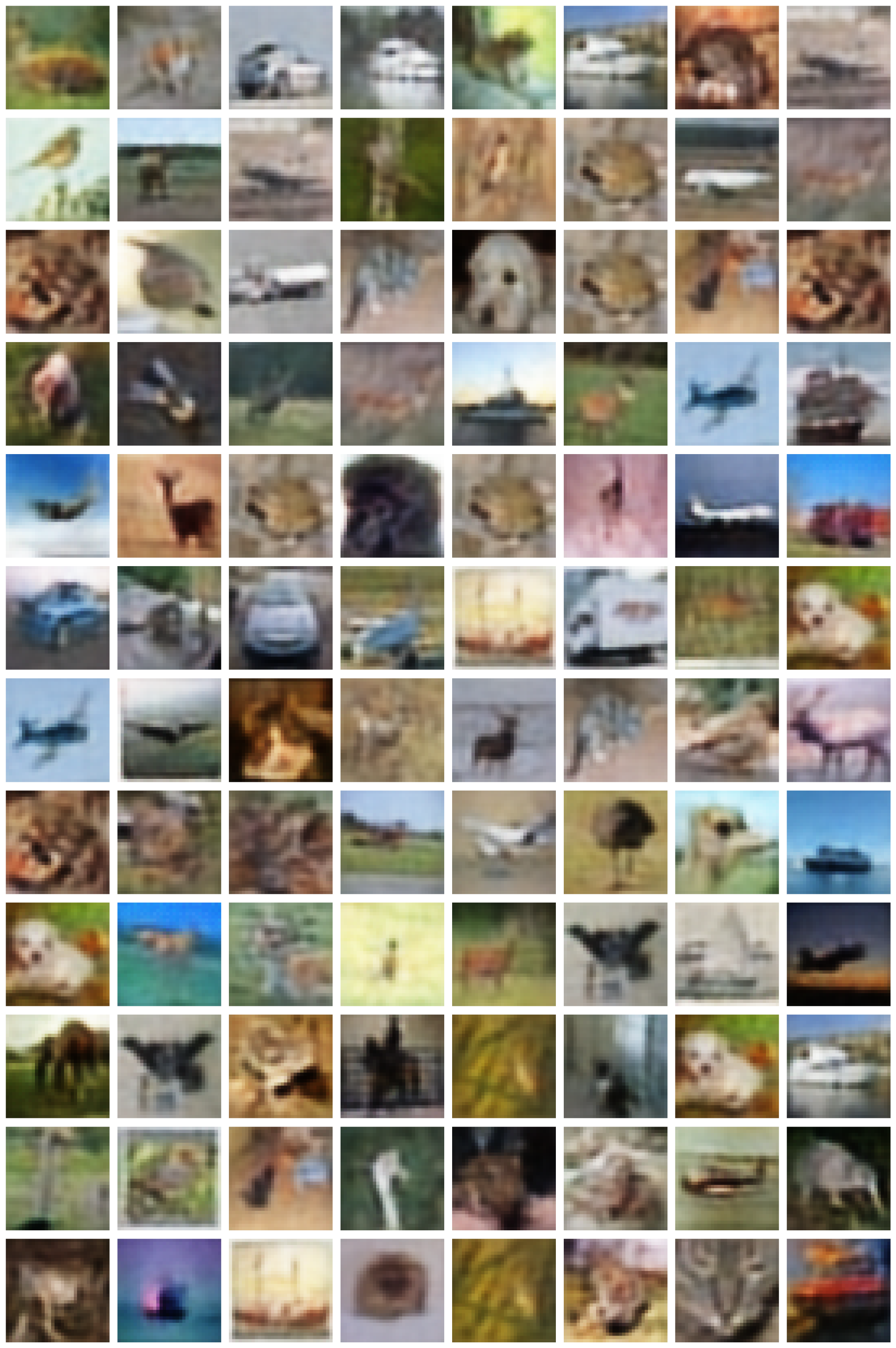}
  \caption{Uncurated generative samples from our Kähler-sampled VAE on our CIFAR-10 experiment with curvature-aware posterior sampling. We choose noise scaling parameter of $1$ here (meaning we take $\text{jitter} \cdot \epsilon \odot \sigma$, $\text{jitter}=1$ for our sampling).}
  \label{fig:cifar_sampling}
\end{figure}

\end{document}